\documentclass[journal]{IEEEtran}
\usepackage{graphicx}
\usepackage{comment}
\usepackage{amsmath,amssymb} 
\usepackage{color}
\usepackage{algorithm}
\usepackage{algorithmic}
\usepackage{epsfig}
\usepackage{graphics}
\usepackage{ulem}
\usepackage{color}
\usepackage{mathrsfs}
\usepackage[table]{xcolor}
\usepackage{booktabs}
\usepackage{makecell}
\usepackage{colortbl}
\usepackage{multirow}
\usepackage{array}
\usepackage{colortbl}
\usepackage{bm}
\usepackage{balance}
\usepackage[switch,columnwise]{lineno}
\IEEEpubid{\begin{minipage}{\textwidth}\ \\[12pt]  \centering \\
  \vspace{1.0cm}
  This article has been accepted for publication in a future issue of this journal, but has not been fully edited.  Content may change prior to final publication. \\ Citation information: DOI 10.1109/TCSVT.2021.3078517, IEEE Transactions on Circuits and Systems for Video Technology. \\
  1051-8215 \copyright 2021 IEEE. Personal use is permitted, but republication/redistribution requires IEEE permission.\\
  See http://www.ieee.org/publications standards/publications/rights/index.html for more information.
\end{minipage}}

\ifCLASSINFOpdf
\else
\fi
\begin{document}
\title{Joint Face Image Restoration and Frontalization for \\ Recognition}

%

\author{Xiaoguang Tu,
        Jian Zhao$^*$,~\IEEEmembership{Member, ~IEEE,
        Qiankun Liu,
        Wenjie Ai, \\
        Guodong Guo,
        Zhifeng Li,
        Wei Liu, ~\IEEEmembership{Senior Member, ~IEEE} and
        Jiashi Feng,~\IEEEmembership{Member},~IEEE}
\thanks{Xiaoguang Tu is with Aviation Engineering Institute, Civil Aviation Flight University of China, Guanghan, China. Email: xguangtu@outlook.com}
\thanks{Wenjie Ai is with School of Information and Communication Engineering, University of Electronic Science and Technology of China, Chengdu, China. Email: 201821011405@std.uestc.edu.cn.}
\thanks{Jian Zhao is with the Institute of North Electronic Equipment, Beijing, China. Homepage: https://zhaoj9014.github.io/. Email: zhaojian90@u.nus.edu.}%
\thanks{Qiankun Liu is with Pensees Ptd Ltd, Singapore. Email: allen.liu@pensees.ai.}%
\thanks{Guodong Guo is with  Institute of Deep Learning, Baidu Research, Beijing, China. Email: guodong.guo@mail.wvu.edu.}%
\thanks{Zhifeng Li and Wei Liu are with Tencent AI Lab, Shenzhen, China. Email: \{michaelzfli@tencent.com, wl2223@columbia.edu.\}}%
\thanks{Jiashi Feng is with National University of Singapore, Singapore. Email: elefjia@nus.edu.sg.}%
\thanks{$^*$ Jian Zhao is the corresponding author.}%
\thanks{Manuscript received ; revised August .}}


\maketitle

\begin{abstract}
In real-world scenarios, many factors may harm face recognition performance, \textit{e.g}., large pose, bad illumination, low resolution, blur and noise. To address these challenges, previous efforts usually first restore the low-quality faces to high-quality ones and then perform face recognition. However, most of these methods are stage-wise, which is sub-optimal and  deviates from the reality. In this paper, we address all these challenges jointly for unconstrained face recognition. We propose an \textbf{M}ulti-\textbf{D}egradation \textbf{F}ace \textbf{R}estoration (MDFR) model to restore frontalized high-quality faces from the given low-quality ones under arbitrary facial poses, with three distinct novelties. First, MDFR is a well-designed encoder-decoder architecture which extracts feature representation from an input face image with arbitrary low-quality factors and restores it to a high-quality counterpart. Second, MDFR introduces a pose residual learning strategy along with a 3D-based \textbf{P}ose \textbf{N}ormalization \textbf{M}odule (PNM), which can perceive the pose gap between the input initial pose and its real-frontal pose to guide the face frontalization. Finally, MDFR can generate frontalized high-quality face images by a single unified network, showing a strong capability of preserving face identity. Qualitative and quantitative experiments on both controlled and in-the-wild benchmarks demonstrate the superiority of MDFR over state-of-the-art methods on both face frontalization and face restoration.
\end{abstract}

\begin{IEEEkeywords}
3D based Face Normalization, Multi-Degradation Face Restoration, Unconstrained Face Recognition.
\end{IEEEkeywords}
%
\IEEEpeerreviewmaketitle

\section{Introduction}
\begin{figure}[!t]
\centering{\includegraphics[width = 8.5cm, height=6.5cm]{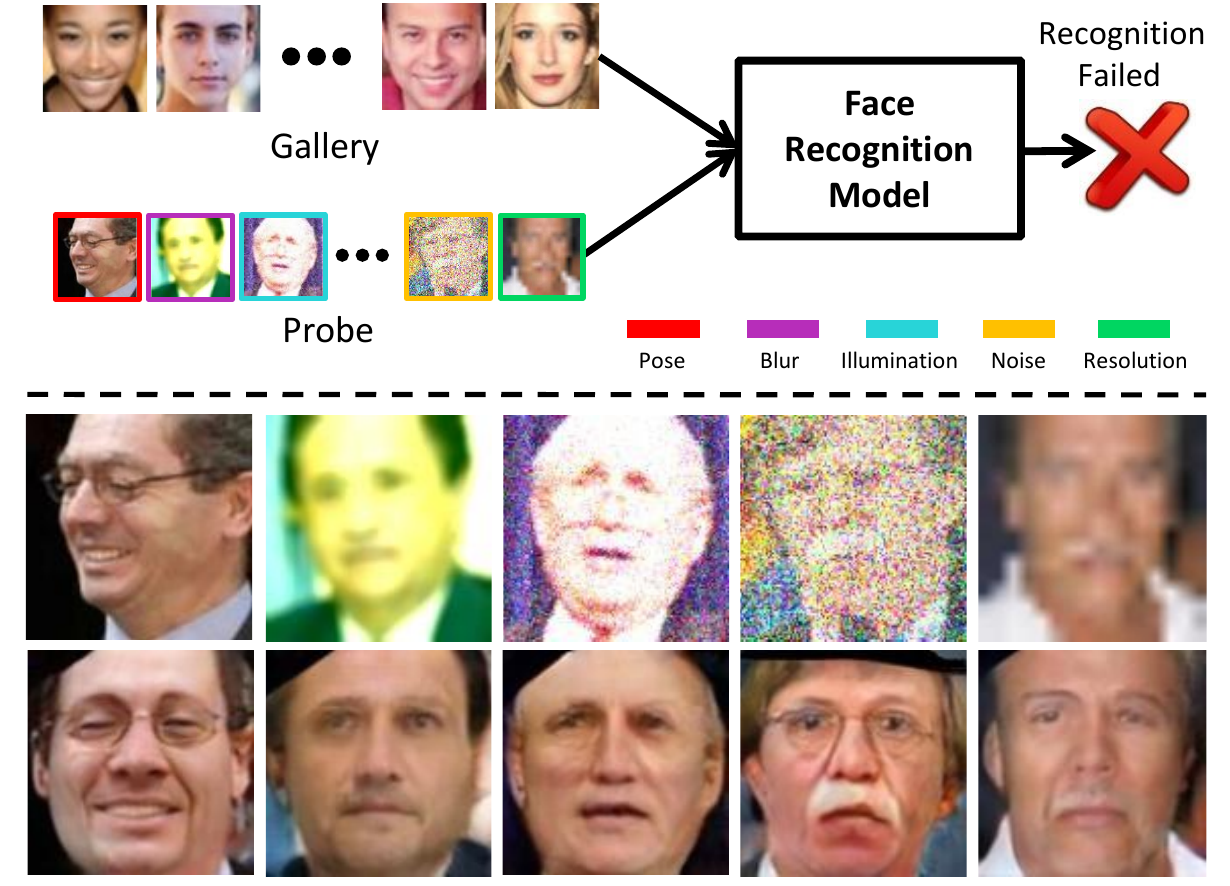}}
\vspace{-0.28cm}
\caption{Top: Illustration of challenging unconstrained face recognition. The presence of large poses and low quality fails the face recognition system. Each color in the legend indicates one influencing factor. Bottom: Example face images recovered by MDFR. \textit{Row} 1: Input profile and low-quality face images (some   images contain multiple contaminating factors). \textit{Row} 2: Recovered frontalized high-quality faces.
\label{fig:1}}
\end{figure}

Unconstrained face recognition \cite{klare2015pushing,zhao20183d,zhao2019multi,masi2019face,zhao2019recognizing,tu2019enhance,tu2020learning} is an important task in computer vision. In real-world applications, the enrolled faces in the gallery are usually frontal high-quality photos, while the probe faces may show large poses, bad illumination, low resolution, blur and noise, which may fail face recognition systems, as shown in Fig.~\ref{fig:1}. For this reason, the unconstrained face recognition is hardly considered to be solved.

Existing methods typically tackle the above challenges individually. For example, some works \cite{hassner2015effective,tran2017disentangled,zhao2018towards,hu2018pose} synthesize a frontalized face from the profile face to achieve face recognition across poses. Though impressive face recognition accuracy has been achieved on lab-environment datasets such as Multi-PIE \cite{gross2010multi}, their performance drops dramatically on benchmarks containing real-world samples such as IJB-C \cite{maze2018iarpa} with considerable low resolution and blurred data.
Apart from the large pose, another challenge that may fail face recognition is the low quality of  probe images, including bad illumination, low resolution, blur and noise. To improve the robustness of face recognition models against various image quality degradation factors, techniques like super-resolution \cite{yu2017hallucinating,zhang2018super,wang2018esrgan,zhang2020copy,gao2020hierarchical}, illumination normalization \cite{du2010adaptive,ramaiah2015illumination,tu2017illumination,tu2017illumination2}, deblurring \cite{shen2018deep,vasiljevic2016examining} and denoising \cite{wang2018devil} have been proposed. However, these methods merely focus on a single low-quality factor and are less effective for tackling the cases involving multiple challenging factors. For instance, super-resolution \cite{chen2018fsrnet,zhang2018super,kong2019cross} methods are fragile for blurred faces.

Inspecting previous works on face frontalization \cite{hassner2015effective,tran2017disentangled,huang2017beyond,hu2018pose,hsu2017fast} or face restoration \cite{wang2018esrgan,zhang2018super,zhang2018ffdnet,zhou2019awgn,tu2017automatic} from low-quality images, we observe a major problem: they are generally limited to just one of the contaminating factors and easily fail on multi-factor cases. Building a unified model to solve the large pose and low-quality problems at one shot is intuitively a straightforward solution, which however is not easy, considering the complex architecture design as well as the deep entanglement of different contaminating factors. To be more specific, a face restoration model takes the raw image as input while a face frontalization model needs additional input channels to encode the facial pose information, which means their architectures are not compatible.
More importantly, the degraded facial details of input low-quality face images may fail at facial landmark detection and thus hinder face frontalization that heavily relies on landmark information. It is thus very crucial to obtain reliable facial landmarks from low-quality faces.

To tackle these challenges, we develop an \textbf{M}ulti-\textbf{D}egradation \textbf{F}ace \textbf{R}estoration (MDFR) model which is mainly driven by two task-specific generators during training, one for face restoration with multiple low-quality factors and the other for face frontalization. However, even when the two separate generators perform well on each specific task, there is a domain gap between their features in the identity metric space, which makes their identity representations inconsistent and hence affects the final face recognition. This identity inconsistency may in turn affect the results of face generation if these tasks are separately performed. To remove such a domain gap, we further propose an Task-Integrated (TI) training scheme to merge the learnings of these two tasks into a single one, enabling all contamination factors to be tackled by a unified network. Moreover, the TI training ensures face images to be frontalized from a single profile face image, \textit{without requesting any priors such as input facial landmarks and target frontal landmarks}. Please note, although the proposed MDFR is able to jointly address face frontalization and face restoration from multiple degradation factors, it can also perform each task separately, such as face frontalization from high-quality inputs, face super-resolution, face deblurring and denoising and so on.

Structurally, our MDFR consists of two main components, \textit{i.e.,} a dual-agent generator and a dual-agent prior-guided discriminator. The dual-agent generator learns to synthesize frontalized high-quality faces from the degraded inputs via two task-specific agents: a \textbf{F}ace \textbf{R}estoration sub-\textbf{N}et (FRN) and a \textbf{F}ace \textbf{F}rontalization sub-\textbf{N}et (FFN). FRN learns to recover facial details from low-quality images while FFN learns to rotate faces by leveraging the given target facial poses. The dual-agent discriminator consists of a \textbf{P}ose \textbf{C}onditioned \textbf{D}iscriminator (PCD) and an \textbf{I}dentity \textbf{C}onditioned \textbf{D}iscriminator (ICD), which are used to criticize the generated face images by referring to prior knowledge, making the outputs satisfy the input requirements. The well-designed dual-agent generator and dual-agent discriminator work together to achieve high-fidelity and identity-preserving frontal face generations from the low-quality inputs. The proposed training scheme is a two-stage training strategy, which contains the separate training and TI training. The 3D-based \textbf{P}ose \textbf{N}ormalization \textbf{M}odule (PNM) is used to guide real-frontal face generation during TI training, which merges face frontalization and restoration into a single unified network, so that the tasks could be blended into the same identity representation space and boost each other to learn more powerful representations for recognition. Our contributions are summarized as follows:

\begin{itemize}
\setlength{\parskip}{-0.80pt}
\item We propose a novel \textbf{M}ulti-\textbf{D}egradation \textbf{F}ace \textbf{R}estoration (MDFR) model which recovers frontalized high-quality face images from given face images with arbitrary poses and multiple low-quality factors.

\item We formulate face frontalization by pose residual learning and propose a 3D-based \textbf{P}ose \textbf{N}ormalization \textbf{M}odule (PNM), which normalizes 2D facial landmarks to real-frontal for guiding face frontalization learning.

\item We develop an effective TI training strategy to merge face restoration and frontalizaiton into a unified network, which further enhances the output quality and improves the face recognition performance.

\item Our method shows the ability to synthesize photorealistic frontal faces from low-quality faces with arbitrary poses and achieves remarkable face recognition performance under unconstrained environments.

\end{itemize}

\section{Related Work}
\sloppy{}
\subsection{Face Frontalization}
Earlier methods address face frontalization through 2D/3D local texture warping \cite{hassner2015effective,zhu2015high} or statistical modeling \cite{sagonas2015robust}. In \cite{kan2014stacked}, Kan \textit{ et al.} use Stacked Progressive Auto-Encoders to rotate a profile face to frontal. Later, Hassner \textit{et al.} \cite{hassner2015effective} apply a single and unmodified 3D surface to approximate the shape of all the input faces. In \cite{sagonas2015robust}, Sagonas \textit{et al.} propose joint frontal face reconstruction and landmark detection by solving a constrained low-rank minimization problem. Such methods show effectiveness on face frontalization, but they tend to suffer a great performance degradation for profile and near-profile\footnote{Faces with yaw angle greater than $60^\circ$.} faces due to severe texture loss and artifacts.

\begin{figure*}[!t]
\centering{\includegraphics[width = 17.5cm, height=8.5cm]{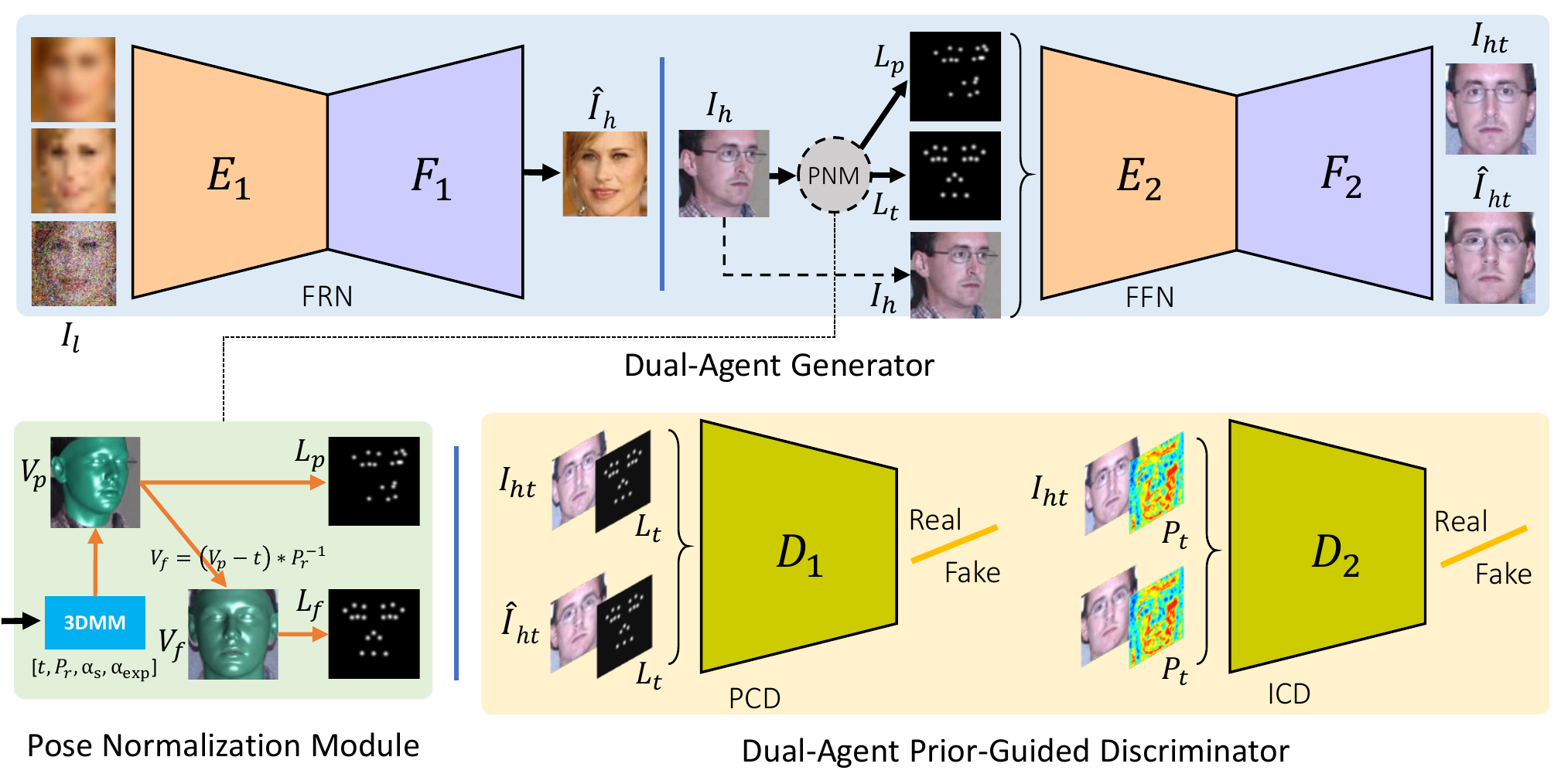}}
\vspace{-0.3cm}
\caption{Overview of MDFR. MDFR consists of two main components, \textit{i.e.,} the dual-agent generator which includes a \textbf{F}ace \textbf{R}estoration sub-\textbf{N}et (FRN) and a \textbf{F}ace \textbf{F}rontalization sub-\textbf{N}et (FFN), and the dual-agent prior-guided discriminator which contains a \textbf{P}ose \textbf{C}onditioned \textbf{D}iscriminator (PCD) and an \textbf{I}dentity \textbf{C}onditioned \textbf{D}iscriminator (ICD). During the task-integrated training, the \textbf{P}ose \textbf{N}ormalization \textbf{M}odule (PNM) is proposed to offer real-frontal facial landmarks $L_f$, serving as the target facial landmarks $L_t$ for guiding face generation
\label{fig:2}}
\end{figure*}

With the advent of GANs \cite{goodfellow2014generative} in the field of computer vision, several GAN-based approaches \cite{yin2017towards,zhao2017dual,huang2017beyond,zhao2018towards} have been proposed to synthesize frontal face images from the profile counterparts. In \cite{tran2017disentangled}, Tran \textit{et al.} propose the DR-GAN for frontal face generation for the first time. Then, TP-GAN \cite{huang2017beyond} is proposed with a two-pathway structure and perceptual supervision. It leverages a well pre-trained face recognition model to guide an identity preserving inference of frontal views from profiles. PIM \cite{zhao2018towards} aims to generate high-quality results through adding regularization terms to learn more robust face representations. In \cite{tian2018cr}, Tian \textit{et al.} introduce a generation sideway to maintain the completeness of the learned embedding space and utilize both labelled and unlabelled data to further enrich the embedding space for realistic generations.

All these methods treat face frontalization as a direct 2D image-to-image translation problem and use frontal faces as ground-truths. However, we argue that such frontal faces are \textbf{not real ground-truths}, say near-frontal, as they may differ from the real world frontal poses at pixel level due to the variations in the collecting process. \textit{If such pseudo ground-truths are used directly for training, the model can hardly converge well}. Even though CAPG-GAN \cite{hu2018pose} can alleviate this issue by offering a target pose, it still needs external assistance to obtain the target frontal pose. In this paper, we formulate face frontalization by pose residual learning and introduce a \textbf{P}ose \textbf{N}ormalization \textbf{M}odule (PNM) based on 3D face morphable model to offer real-frontal poses for face frontalization. PNM projects facial landmarks into a standard 3D space and automatically rotates them to real-frontal, serving as a pose target for refining the outputs towards frontal restoration. Once the training is done, our model can generate real-frontal face images from a single input \textbf{without requesting any target poses}.

\subsection{Face Restoration}
To restore high-quality face images from low-quality counterparts, methods such as super-resolution \cite{kim2016accurate,lai2017deep,chen2018fsrnet}, denoising \cite{mosleh2014image,yue2014cid}, deblurring \cite{anwar2015class,svoboda2016cnn,xu2017learning} and illumination normalization \cite{vishwakarma2015illumination,yu2017discriminative,wang2011illumination} have been proposed. For example, Kim \textit{et al.} \cite{kim2016accurate} utilize a very deep convolutional network by cascading many small filters to extract contextual information for super-resolution. Lai \textit{et al.} \cite{lai2017deep} propose the LPSR Network to restore high-resolution images based on cascaded CNNs.  In \cite{chen2018fsrnet,yu2018face,yu2018super}, the researchers recover high-resolution images with the aid of facial attributes, \textit{i.e.}, facial landmark, parsing segmentation information and geometry prior estimation. Besides, some works \cite{yu2020hallucinating,zhang2021recursive,yu2018imagining} focus on architecture design to improve the super-resolution performance. For image deblurring or denoising, earlier methods \cite{mosleh2014image,yue2014cid,anwar2015class} mainly exploit frequency-domain knowledge to restore band-pass frequency components. In \cite{svoboda2016cnn}, Svoboda \textit{et al.} tackle this problem using custom CNN for the first time. Later, Xu \textit{et al.} \cite{xu2017learning} perform joint deblurring and super-resolution for face and text images with GANs to learn a category-specific prior to solve this problem. In \cite{shen2018deep}, Shen \textit{et al.} exploit global and local semantic cues and incorporate perceptual and adversarial losses to restore photorealistic face images with finer details. Illumination mainly changes the weight of pixel values in a face image, which also leads to degraded recognition performance in extreme conditions. Prior methods addressing this problem are mainly based on holistic normalization \cite{vishwakarma2015illumination,yu2017discriminative} or invariant feature extraction \cite{wang2011illumination,ramaiah2015illumination}. Methods of the first category redistribute the intensities of the original image in a more normalized way, which is less prone to lighting changes; invariant feature extraction methods extract illumination-invariant features, such as the high-frequency and gradient-based components.

Though the mentioned methods are effective for face image enhancement, their performances usually drop when encountering cross degradation factors. For instance, using the deblurring to enhance low-resolution images would not help, and sometimes leads to worse performance because of overfitting to blurring factors.

\section{Multi-Degradation Face Restoration}
An overview of the proposed MDFR is shown in Fig.~\ref{fig:2}. As can be seen, our model consists of a Dual-Agent Generator, a Dual-Agent Prior-Guided Discriminator and a Pose Normalization Module. We now present each component in details.

\subsection{Dual-Agent Generator}

\begin{figure*}[!t]
\centering{\includegraphics[width = 17.5cm, height=6.2cm]{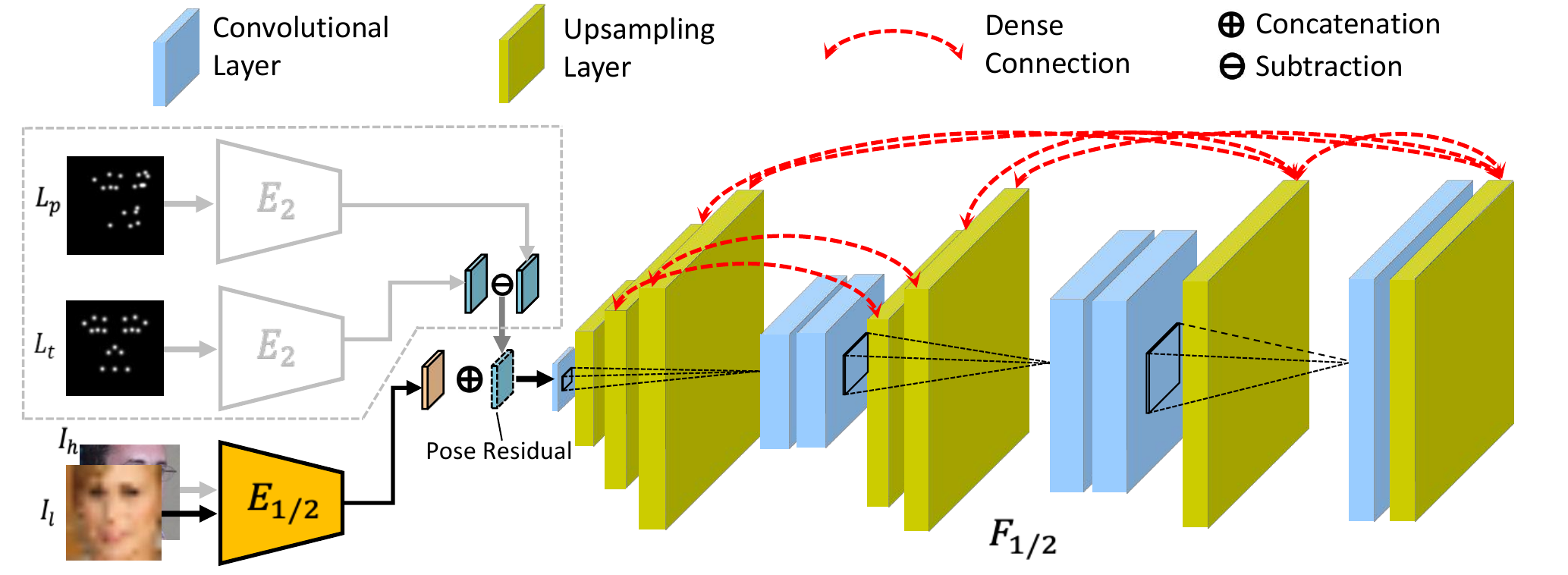}}
\vspace{-0.2cm}
\caption{Architecture of FRN/FFN. FRN and FFN share the same architecture. Decoder $F_1$ of FRN takes as input the representations of the input image encoded by $E_1$, while $F_2$ of FFN takes as input the concatenated representation of the input image and the pose residual between profile and frontal facial landmarks
\label{fig:3}}
\end{figure*}

The dual-agent generator contains a \textbf{F}ace \textbf{R}estoration sub-\textbf{N}et (FRN) and a \textbf{F}ace \textbf{F}rontali-zation sub-\textbf{N}et (FFN), each consisting of an encoder to map the input into an embedding space, and a decoder to recover the embedding code to the target face, with the same architecture but taking in different inputs, as shown in Fig.~\ref{fig:3}.

FRN takes as input a low-quality face $I_l$ and outputs a high-quality counterpart $\hat{I}_h$:
\begin{equation}
\small
\begin{aligned}
\hat{I}_h = G_1(I_l) = F_{1}(E_{1}(I_l)),
\end{aligned}
\end{equation}
where $E_1$ and $F_1$ are the encoder and decoder of FRN, respectively.

FFN takes in three inputs, including a high-quality  face image $I_{h}$, $I_{h}$'s corresponding facial landmarks $L_p$ and a target facial landmarks $L_t$. During the FFN separate training, $L_t$ is the corresponding landmarks of the target near-frontal face image, while during the TI training\footnote{For the details of separate training and task-integrated training, please refer to Sec. 3.4.}, $L_t$ is the real-frontal facial landmarks $L_f$ that normalized by PNM. We use 18 facial landmarks to indicate a facial pose and encode them as Gaussian heatmaps to represent $L_p$ and $L_t$. Different from previous work \cite{hu2018pose} that uses the target pose to directly guide face generation, we feed $L_p$ and $L_t$ into the encoder and perform subtraction to obtain the pose residual. The intuition is that learning only the differences between poses avoids the redundant pose-irrelevant information, such as static backgrounds, which remains unchanged during the transformation. Therefore, the rotated face $\hat{I}_{ht}$ can be generated from $I_{h}$, conditioned on the pose residual:
\begin{equation}
\small
\begin{aligned}
\hat{I}_{ht} = G_2(I_{h}) = F_{2}(E_2(I_{h}) \small{\oplus}\lbrack E_{2}(L_p) - E_{2}(L_t) \rbrack),
\end{aligned}
\end{equation}
where $E_2$ and $F_2$ are the encoder and decoder of FFN, respectively, and $\small{\oplus}$ denotes concatenation. To make the decoder easier reuse features of different spatial positions and facilitate feature propagation, we add dense connections in the decoder. The outputs of each block are connected to the first convolutional layers located in all subsequent blocks in the decoder. As the blocks have different feature resolutions, we upsample feature maps with lower resolutions when we use them as inputs into higher resolution layers.


\subsection{Pose Normalization Module}

We devise a \textbf{P}ose \textbf{N}ormalization \textbf{M}odule (PNM) to perform pose normalization. Note that PNM is only used during the TI training. PNM offers a real-frontal pose with uniform face scale to guide face frontalization. Based on the 3D morphable model \cite{zhu2016face}, the 3D vertices of a 2D face image can be expressed as a linear combination over a set of PCA bases as follows:
\begin{equation}
\small
\begin{aligned}
\bm{S} =  \overline{\bm{S}} + \bm{A}_{\text{s}}\bm{\alpha}_{\text{s}} + \bm{A}_{\text{exp}}\bm{\alpha}_{\text{exp}},
\end{aligned}
\end{equation}
where $\overline{\bm{S}} \in \mathbb{R}^{3 \times {N}}$ is the mean shape, $\bm{A}_\text{s} \in \mathbb{R}^{3 \times {N}}$ is the shape principle basis {trained on the 3D face scans}, $\bm{\alpha}_\text{s} \in \mathbb{R}^{40 }$ is the shape representation coefficient, $\bm{A}_{\text{exp}} \in \mathbb{R}^{3 \times {N}}$ is the expression principle basis, $\bm{\alpha}_{\text{exp}} \in \mathbb{R}^{10}$ is the corresponding expression coefficient, and $N$ is the number of vertices.

The 3D face vertices $\bm{S}$ can be projected onto a 2D image plane with scale orthographic projection to generate the 2D profile face from a specified viewpoint:
\begin{equation}
\small
\begin{aligned}
\bm{V}_p =  f*\bm{Pr}*\bm{\Pi}* \bm{S} + \bm{t},
\end{aligned}
\end{equation}
where $\bm{V}_p$ denotes the 2D coordinates of the 3D vertices projected onto the 2D plane, $f$ is the scale factor, $\bm{\Pi}$ is a fixed orthographic projection matrix, $\bm{Pr}$ is the rotation matrix, and $\bm{t}$ is the translation vector. As $\bm{Pr}$ and $\bm{t}$ indicate the rotation and offset variance, when removed from Eqn. (4), the normative frontal coordinates of a face image with arbitrary poses can be obtained by
\begin{equation}
\small
\begin{aligned}
\bm{V}_f =  f*\bm{\Pi}* \bm{S} = (\bm{V}_p-\bm{t})*Pr^{-1}.
\end{aligned}
\end{equation}
Here $\bm{V}_p$ and $\bm{V}_f$ store the profile and real-frontal dense 2D coordinates $(x,y)$ for a given face image in a standard 3D space, with $z$ coordinates removed. We use the state-of-the-art 3D face reconstruction method 2DASL \cite{tu2019joint,tu20203d} for 3DMM parameters regression and obtain dense coordinates (over 50, 000 points) from a given 2D face image. The 18 common keypoints are sampled from $V_p$ and $V_f$, respectively, to generate the Gaussian heatmaps $L_p$ and $L_f$.

\subsection{Dual-Agent Prior-Guided Discriminator}

The discriminative loss for face super-resolution is firstly proposed in the work URDGN \cite{yu2016ultra}. Following this idea, we propose to condition the discriminator with two kinds of additional prior knowledge, \textit{i.e.,} the target facial landmarks and the frontal face identity feature map, letting the generated image approach the real one not only in terms of the target pose but also in terms of identity representation.

Our prior-guided discriminator is initialized with a VGG-11 \cite{simonyan2014very} backbone. The first discriminator \textbf{P}ose \textbf{C}onditioned \textbf{D}iscriminator (PCD) takes the target pose $L_t$ as condition and pairs with FFN's output $\hat{I}_{ht}$ (or the target high-quality face image $I_{ht}$), \textit{i.e.,} [$\hat{I}_{ht}$, $L_t$] vs. [$I_{ht}$, $L_t$]. The second discriminator \textbf{I}dentity \textbf{C}onditioned \textbf{D}iscriminator (ICD) takes the target face identity feature $P_t$ as condition and pairs with $\hat{I}_{ht}$ or $I_{ht}$, \textit{i.e.},  [$\hat{I}_{ht}$, $P_t$] vs. [$I_{ht}$, $P_t$]. Following such conditions, $F_2$ would generate images $\hat{I}_{ht}$ which approach $I_{ht}$'s appearance and meanwhile satisfy the frontal-pose requirement. Specifically, PCD and ICD can not only distinguish real/fake of the outputs, but also learn the distinctions of facial pose and identity representation between the fake and real images.

\subsection{Overall Training}

Our overall training includes two phases: separate training and TI training. We now describe each training process in details. Algorithm~\ref{alg:A} describes the whole process of the proposed training strategy.

\begin{algorithm}
\caption{Overall Training}
\label{alg:A}
\begin{algorithmic}
\STATE {\textbf{Separate training:} \\
\textbf{Phase 1}: Train face restoration net FRN by $\mathcal{L}_{\text{FRN}}$}, image pair \{$I_l, I_h$\}. $I_l$ is the input low-quality face, $I_h$ is $I_l$'s high-quality face taken as the ground-truth.\\
\textbf{Phase 2}: Train face frontalization net FFN by $\mathcal{L}_{\text{FFN}}$, image pair \{$I_{h}, I_{ht}$\} and landmark pair \{$L_{h}, L_{t}$\}. $I_h$ is the high-quality face, $I_{ht}$ is $I_h$'s target face image (near frontal) taken as ground-truth, \{$L_{h}, L_{t}$\} are their corresponding facial
landmarks.
\STATE {\textbf{Task-Integrated training:} \\
Fixing FFN's parameters. \\
Taking high-quality face $I_h$ as image input of FFN, $I_h$'s real frontal landmark $L_f$ (normalized by PNM) as landmark input of FFN, generating high-quality and frontal face $I_{hf}$. Train face restoration net FRN by $\mathcal{L}_{\text{TI}}$, image pair \{$I_l, I_{hf}$\}. $I_l$ is the low-quality face, $I_{hf}$ is the output of FFN when taking $I_h$ as input. $I_h$ is $I_l$'s corresponding high-quality face image. $I_{hf}$ is used as the ground-truth for FRN training.} \\

1: \textbf{while} not converge \textbf{do} \\
2: \hspace{0.0cm} Choose one minibatch of $N$ low-quality image $I_l^i$, \\
   \hspace{0.4cm} $i=1,...,N$.\\
3: \hspace{0.0cm} FRN outputs one minibatch of $N$ restored images  $\hat{I}_{hf}^i$\\
   \hspace{0.4cm} from the low-quality images $I^i_l$; \\
   \hspace{0.4cm} FFN outputs one minibatch of $N$ images $I_{hf}^i$ from $I_{h}^i$; \\
   \hspace{0.4cm} from the restored images $G_1(I^i_l)$;  \\
4: \hspace{0.0cm} Update FRN by descending its stochastic gradient: \\
   \hspace{0.45cm}$\nabla_{\theta_{FRN}}\frac{1}{N} \sum_{i=1}^{N} \mathcal{L}_{\text{TI}}$. \\
5: \textbf{end while}
\end{algorithmic}
\end{algorithm}

\subsubsection{Separate Training}
We first train FRN and FFN separately. \textbf{FRN} \textbf{S}eparate (FRN-S) training restores high-quality face images from low-quality ones and \textbf{FFN} \textbf{S}eparate (FFN-S) training rotates profile faces to the target pose.

\paragraph{FRN-S Training} The identity-preserving loss $\mathcal{L}_{\text{\text{id}}}$ is employed during FRN-S training to preserve the identity information of the generated face image. We use a pre-trained face recognition model $R_{\text{id}}$ to extract identity features and fix the parameters during training. $\mathcal{L}_{\text{\text{id}}}$ is defined as
\begin{equation}
\small
\begin{aligned}
\mathcal{L}_{\text{\text{id}}}(X, Y) = \left\|\frac{R_{\text{id}}(X)}{||R_{\text{id}}(X)||_2} - \frac{R_{\text{id}}(Y)}{||R_{\text{id}}(Y)||_2}\right\|_2^2,
\end{aligned}
\end{equation}
where $X$ is the input and $Y$ is the output of FRN.

The restoration loss for FRN-S training is defined as
\begin{equation}
\small
\begin{aligned}
\mathcal{L}_{\text{r}}(I_h, \hat{I}_h) = ||I_h - \hat{I}_h||_2^2,
\end{aligned}
\end{equation}
where $\hat{I}_h$ is the restored face image by FRN-S and $I_h$ is the high-quality face image. The overall loss function for FRN-S is
\begin{equation}
\small
\begin{aligned}
\mathcal{L}_{\text{FRN}}(I_h, \hat{I}_h) = \mathcal{L}_{\text{r}}(I_h, \hat{I}_h) + \lambda_1 \mathcal{L}_{\text{id}}(I_h, \hat{I}_h),
\end{aligned}
\end{equation}
where $\lambda_1$ is a weighting parameter balancing different losses.

\paragraph{FFN-S Training} The FFN-S training is supervised by four losses, \textit{i.e.}, identity-preserving loss $\mathcal{L}_{\text{id}}$, frontalization loss $\mathcal{L}_{\text{f}}$, and conditional adversarial losses $\mathcal{L}_{\text{pcd}}$ and $\mathcal{L}_{\text{icd}}$. $\mathcal{L}_{\text{id}}$ for FFN-S is the same as that for FRN-S. For $\mathcal{L}_{\text{f}}$, we penalize the pixel-wise Euclidean distance between the rotated image and its corresponding ground-truth,
\begin{equation}
\small
\begin{aligned}
\mathcal{L}_{\text{f}}(I_{ht}, \hat{I}_{ht}) = ||I_{ht} - \hat{I}_{ht}||_2^{2},
\end{aligned}
\end{equation}
where $I_{ht}$ is the target near-frontal face image and $\hat{I}_{ht}$ is the output of FFN.

The decoder of FFN takes as input the latent code of the profile face and the pose residual between the profile facial landmark heatmap $L_p$ and the target facial landmark heatmap $L_t$, which provides strong guidance for face rotation. The dual-agent discriminator is used to refine $\hat{I}_{ht}$ according to the given prior knowledge. For PCD, $D_1$ takes $L_t$ as condition and pairs with $\hat{I}_{ht}$ and $I_{ht}$ as input. For ICD, $D_2$ takes the identity feature map $P_t$ as condition and pairs with $\hat{I}_{ht}$ and $I_{ht}$ as input. The condition adversarial losses $\mathcal{L}_{\text{pcd}}$ and $\mathcal{L}_{\text{icd}}$ can thus be defined as
\begin{equation}
\small
\begin{aligned}
\mathcal{L}_{\text{pcd}} &=  \mathbb{E}_{I_p\in\mathcal{I}}[\log(D_1([L_t, I_{ht}])) + \log(1-D_1([L_t, \hat{I}_{ht}]))],  \\
\mathcal{L}_{\text{icd}} &=  \mathbb{E}_{I_p\in\mathcal{I}}[\log(D_2([P_t, I_{ht}])) + \log(1-D_2([P_t, \hat{I}_{ht}]))].
\end{aligned}
\end{equation}
The overall loss function is a weighted sum of the above losses. The parameters of generator ($\theta_G$), PCD ($\theta_{P}$) and ICD ($\theta_{I}$) are trained alternatively to optimize the following min-max problem:
\begin{equation}\label{11}
\small
\min\limits_{\theta_G} \max\limits_{\theta_{P}, \theta_{I}} \mathcal{L}_{\text{FFN}} = \mathcal{L}_{\text{f}}(I_{ht}, \hat{I}_{ht}) + \lambda_2 \mathcal{L}_{\text{id}}(I_{ht}, \hat{I}_{ht}) + \lambda_3( \mathcal{L}_{\text{pcd}} + \mathcal{L}_{\text{icd}}),
\end{equation}
where $\lambda_2$ and $\lambda_3$ are weighting parameters trading off different losses.

\subsubsection{Task-Integrated (TI) Training}
After FRN and FFN are pre-trained, we perform the TI training, which we term as FRN Task-Integrated (FRN-TI) training. We use the output of FFN as ground-truth to train FRN. FRN-TI behaves somewhat like the distillation \cite{gupta2016cross} process by transferring knowledge from the teacher model to the student. During FFN-S training, FFN learns to generate the target face according to the given target facial landmarks. During TI training, we use the real-frontal facial landmarks $L_f$ that normalized by PNM to guide face generation for FFN, so the normalized pose representation is embedded into the outputs as well as the feature maps of FFN, which can serve as ground-truth for face frontalization by FRN. We thus formulate the training of FRN-TI at both image and feature levels. Specifically, we use FFN's output as well as its deep feature maps as the ground-truth to guide the learning of FRN. After the training is completed, PNM and FFN could be removed and the high-quality frontal face can be generated by merely using FRN without given the target facial landmarks. We train FRN-TI end-to-end. During the FRN-TI training, the parameters of FFN are fixed and only FRN is optimized.

We use the last dense block of FRN and FFN to perform feature level supervision via a \textbf{F}eature \textbf{A}lignment (FA) loss. As FRN and FFN have the same architecture, the FA loss $\mathcal{L}_{\text{FA}}$ can be easily defined as the mean squared error between their feature maps:
\begin{equation}
\small
\begin{aligned}
\mathcal{L}_{\text{FA}} = \frac{1}{N} \left\|\sum_{i=0}^{N}(B^i_{FFN}(G_1(I_l)) - B^i_{FRN}(I_l)) \right\|_2^2,
\end{aligned}
\end{equation}
where $I_l$ is an arbitrary low-quality face image, $B_{FRN}(\cdot)$ and $B_{FFN}(\cdot)$ are the feature representations from the last block of the decoders $F_1$ and $F_2$, respectively, $N$ is the number of feature maps, and $i$ is the $i$-th feature map.

Then, the overall loss function for TI training is
\begin{equation}
\small
\begin{aligned}
\mathcal{L}_{\text{TI}} = \mathcal{L}_{\text{r}}(\hat{I}_{ht}, G_1(I_l)) + \lambda_4\mathcal{L}_{\text{id}}(\hat{I}_{ht}, G_1(I_l)) + \lambda_5 \mathcal{L}_{\text{FA}},
\end{aligned}
\end{equation}
where $\hat{I}_{ht}$ is the output of FFN, and $\lambda_4$ and $\lambda_5$ are weighting parameters among different losses. During FRN-TI training, FFN generates aligned images and features to guide face generation in FRN. After the task-integrated training is completed, FRN-TI is capable of generating frontalized high-quality faces by itself.

\section{Experiments}

\textbf{Implementation} The size of face images is fixed as $128 \times 128$; constraint factors $\lambda_1$, $\lambda_2$, $\lambda_3$, $\lambda_4$, and $\lambda_5$ are fixed as $10^4$, $10^4$, $10^4$, $0.1$ and 1, respectively; batch size is set as 8; initial learning rate $lr$ for FRN, FFN, PCD and ICD is $10^{-4}$, $10^{-4}$, $10^{-3}$ and $10^{-3}$, respectively. We use 2DASL \cite{tu2019joint} for 3D face reconstruction. We initialize the face recognition network with ResNet-50 \cite{he2016deep} to extract face identity features, which is pre-trained on CASIA-Webface \cite{yi2014learning} dataset using the AAM loss \cite{deng2019arcface}.

\subsection{Datasets}

\paragraph{CASIA-Webface} The CASIA-Webface \cite{yi2014learning} dataset contains 494,414 face images from 10,575 identities detected from the Internet. We use it for face recognition model training and FRN-S training. To generate the low-quality face images, we use color warp to randomly change the pixel's RGB value; Gaussian, uniform and average filters to randomly perform blurring; Gamma adjust to randomly change images' brightness level; bicubic to perform down-sampling and Gaussian noise to contaminate the images. The mean and standard deviation for Gaussian filter is a random number ranged from [0.1, 0.2] and [0.1, 0.2] , respectively. The mean and standard deviation for color wrap is a random number ranged from [0.1, 0.2] and [0.1, 0.2] , respectively. The low and high bound for Gamma adjust is a random number ranged from [0.1, 0.3] and [1, 3], respectively. The mean and standard deviation for Gaussian noise is a random number ranged from [0.1, 0.5] and [0.1, 0.5], respectively.

\paragraph{Multi-PIE} The CMU Multi-PIE \cite{yi2014learning} is the largest multi-view face recognition benchmark and is collected in four sessions. Following previous face frontalization works \cite{zhao2018towards,hu2018pose}, we conduct experiments under two settings: \textbf{Setting-1} only uses the images in session one, which contains 250 identities. The images with 11 poses within 90$^\circ$ of the first 150 identities are used for training. For testing, one frontal view with neutral expression and illumination is used as the gallery image for each of the remaining 100 identities and other images are used as probes. \textbf{Setting-2} uses the images with neutral expression from all four sessions, which contains 337 identities. The images with 11 poses within 90$\circ$ of the first 200 identities are used for training. For testing, one frontal view with neural illumination is used as the gallery image for each of the remaining 137 identities and other images are used as probes. The training subsets of Multi-PIE are used for FFN-S training and FRN-TI training. The testing subsets are used for face frontalization evaluation.

\paragraph{LFW} The Labelled Faces in the Wild (LFW) \cite{huang2008labeled} dataset contains 13,233 high-quality face images of 5,749 identities. The images are obtained by trawling the Internet followed by face centering, scaling and cropping based on bounding boxes provided by an automatic face locator. We use LFW for face frontalization and restoration testing.

\paragraph{IJB-C} The IARPA Janus Benchmark (IJB-C) \cite{maze2018iarpa} contains both still images and video frames from ``in-the-wild'' environment, which is believed to be the most unconstrained face dataset to date. We use it for face restoration testing.

\paragraph{CelebA} CelebFaces Attributes Dataset (CelebA) is a large-scale face attribute dataset with more than 200K celebrity images, each with 40 attribute annotations. The images in this dataset cover large pose variations and background clutter. We use this dataset for model testing.

\subsection{Evaluation on Face Frontalization}

We first verify MDFR's effectiveness on face frontalization. During inference, we feed FFN-S and FRN-TI with the high-quality profile images, thus FRN-TI can be viewed as solely focusing on face frontalization.

\begin{figure*}[!t]
\centering{\includegraphics[width = 17.8cm, height=8.5cm]{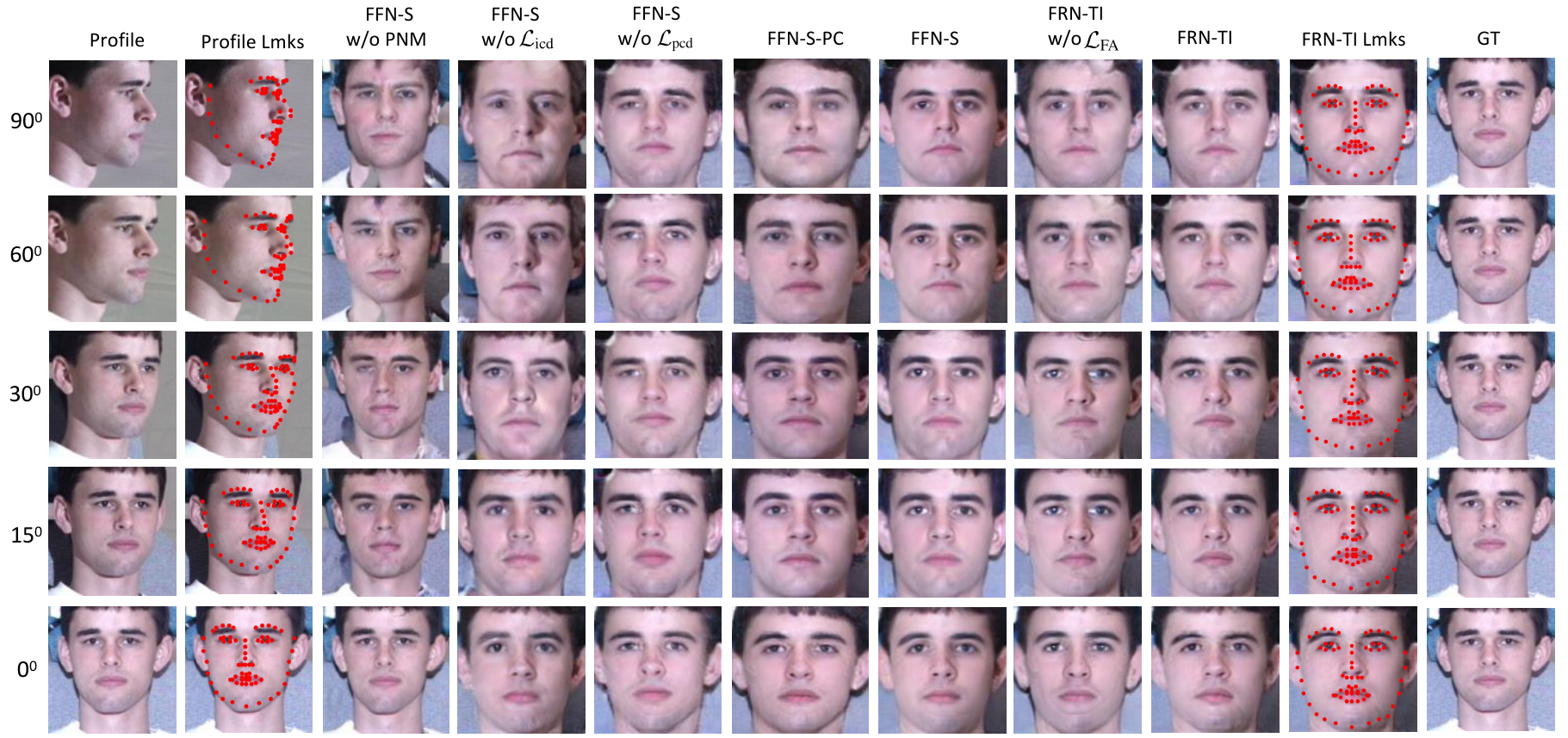}}
\vspace{-0.2cm}
\caption{Component analysis. Outputs of FFN-S and FRN-TI and their variants. \textit{Col.} 2 and \textit{Col.} 10 are the landmark detection results by FAN \cite{bulat2017far} for the profile images and FRN-TI outputs, respectively.
\label{fig:4}}
\end{figure*}

\begin{table*}
\small
\begin{center}
\caption{Component analysis. Rank-1 recognition rates (\%) under Multi-PIE [12] Setting-1. b denotes the performance of ResNet-50 on original profile images. }
\resizebox{0.6 \textwidth}{!}{
\begin{tabular}{c l c c c c c c}
\hline
1 & Methods \quad & $\pm 90^o$ & $\pm 75^o$ & $\pm 60^o$ & $\pm 45^o$  & $\pm 30^o$ & $\pm 15^o$\\
\hline
2 & b & 18.80 & 63.82 & 92.21 & 98.30 & 99.23 & 99.40 \\
\hline
3 & FFN-S w/o PNM & 63.27 & 81.32 & 91.05 & 97.12 & 98.40 & 98.54\\
4 & FFN-S w/o $\mathcal{L}_{\text{icd}}$ & 43.53 & 66.34 & 76.50 & 84.21 & 92.62 & 93.56\\
4 & FFN-S w/o $\mathcal{L}_{\text{pcd}}$ & 63.70 & 82.04 & 91.41 & 97.42 & 98.58 & 98.54\\
 \rowcolor{gray!20}
6 & FFN-S & 80.11 & 89.27 & 94.94 & 99.06 & 99.51 & 99.92\\
7 & FRN-TI w/o $\mathcal{L}_{\text{fa}}$ & 75.47 & 86.83 & 93.17 & 98.12 & 98.92 & 99.68\\ \rowcolor{gray!20}
8 & FRN-TI & 79.83 & 88.52 & 94.20 & 98.78 & 99.36 & 99.90\\
\hline
\end{tabular}}
\end{center}
\label{tab:0}
\end{table*}

\begin{figure*}[!t]
\hspace{-0.5cm}
\centering{\includegraphics[width = 17.5cm, height=11.3cm]{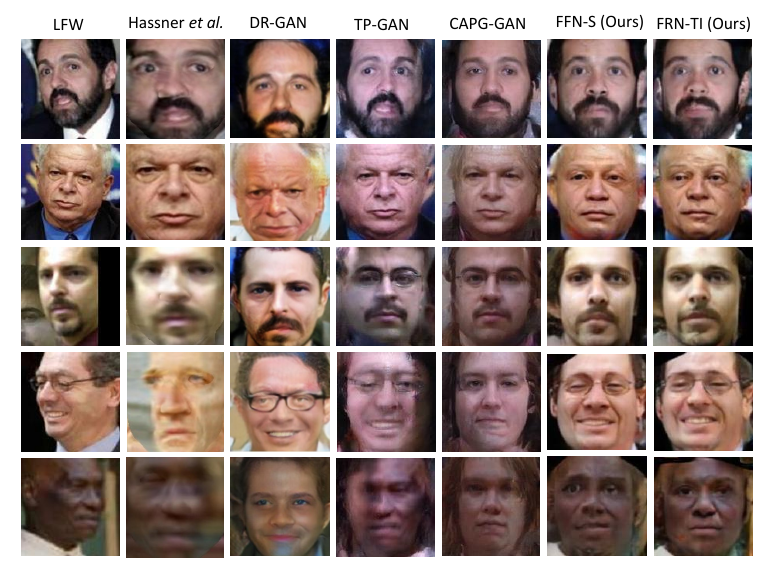}}
\caption{The comparison results between our model and other popular face frontalization methods on LFW. The inputs to FFN are high-quality face images with arbitrary facial poses.
\label{fig:5}}
\end{figure*}

\begin{figure*}[!t]
\hspace{-0.5cm}
\centering{\includegraphics[width = 17.5cm, height=5.2cm]{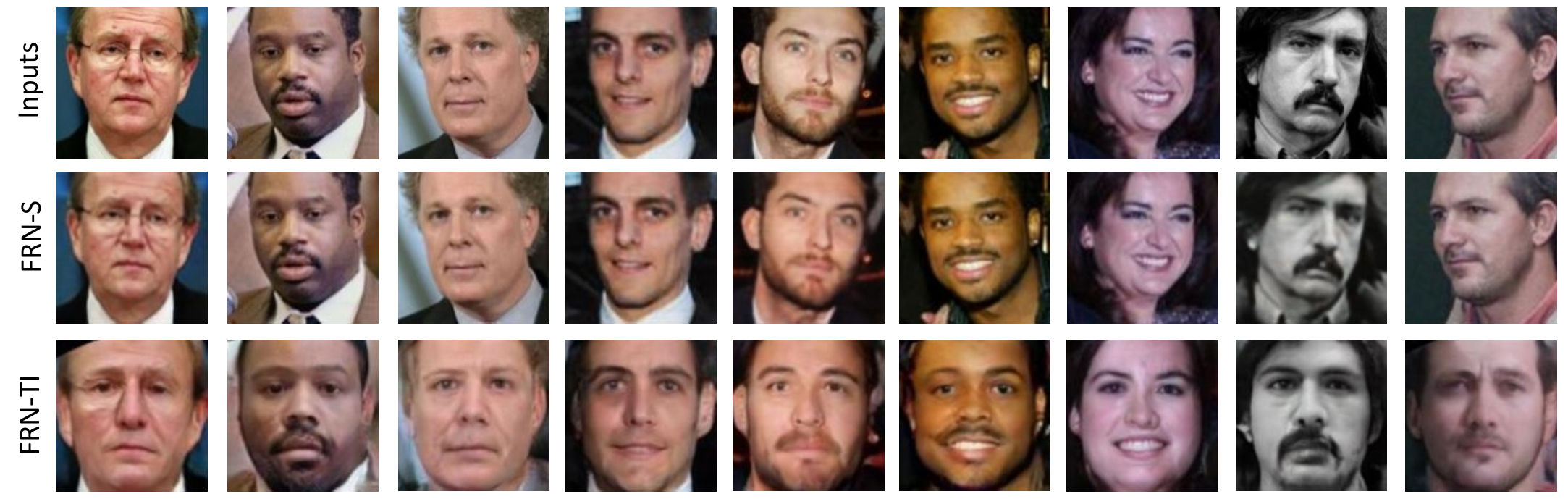}}
\caption{The results of FRN-S and FRN-TI on LFW, respectively. The images of the first row are the high-quality inputs, the images of the second row are the outputs of FRN-S, the images of the third row are the outputs of FRN-TI.
\label{fig:add1}}
\end{figure*}

\subsubsection{Component Analysis}
We investigate different architectures and loss combinations of FFN-S and FRN-TI to see their respective roles in face generation. We compare the results of FFN-S with three variants: w/o Pose Normalization Module (PNM), w/o $\mathcal{L}_{\text{pcd}}$ and w/o $\mathcal{L}_{\text{icd}}$, in each case. For FFN-S w/o PNM, we remove the facial landmark heatmaps from both FFN's encoder and PCD. We also compare the results with FRN-TI by using/removing $\mathcal{L}_{\text{FA}}$ to investigate $\mathcal{L}_{\text{FA}}$'s effectiveness when performing FRN-TI training. To evaluate the effectiveness of the pose residual learning module, we compare FFN-S with the modified one, which directly concatenates the pose features together. The modified pose learning strategy is named FFN-S-PC (Pose Concatenating).

The visualized results of all variants are shown in Fig.~\ref{fig:4}, where we observe all the variants perform well within a pose range of $\pm 30^\circ$. However, when the pose is larger than $30^\circ$, FFN-S w/o PNM, $\mathcal{L}_{\text{icd}}$ and $\mathcal{L}_{\text{pcd}}$ all arise artifacts to some extent. PNM offers real-frontal landmarks to guide face frontalization. If removed, the model is unable to locate face regions correctly, making the recovered face distorted in part of face regions (see \textit{Col.} 3). As $\mathcal{L}_{\text{icd}}$ helps refine the generated face images according to the given identity information, if removed, the output's identities are totally changed in larger poses (\textit{Col. 4}). Although the identity-preserving loss $\mathcal{L}_{\text{id}}$ draws the input and output closer in the identity metric space, it is difficult to affect the appearance at image level. However, $\mathcal{L}_{\text{icd}}$ is able to directly refine the output to its original appearance by distinguishing real/fake between the concatenation pair of face image and identity feature map. For FFN-S w/o $\mathcal{L}_{\text{pcd}}$, the outputs (\textit{Col.} 5) are not real-frontal but slightly rightward. This small angle drift may make the outputs involve some artifacts for large pose frontalization. FRN-TI achieves nearly the same visual results with FFN-S, which verifies the effectiveness of our proposed TI training. However, if $\mathcal{L}_{\text{FA}}$ is removed during FRN-TI training, the visual performance drops slightly, \textit{e.g.,} the eyes that look strange in \textit{Col.} 8. For FFN-S-PC (\textit{Col}. 6), it achieves comparable results with FFN-S within small pose changes ($\leq 45^\circ$). However for the poses larger than $45^\circ$, especial for the pose of $90^\circ$, the result of FFN-S-PC is not as clear as that of FFN-S. The reason should be that the concatenation of pose heatmaps can not well eliminate the background, which may influence the generation process, causing the degradation of the generated results.

The averaged rank-1 recognition rates of all variations are compared on \textbf{Setting-1} in Tab.~\uppercase\expandafter{\romannumeral1}. The results on the original profile images serve as our baseline (\textit{i.e.}, b).  The results of all experiments are based on Resnet-50 \cite{he2016deep} for face recognition. By comparing the results of \textit{row} 2-8, we observe FFN-S and FRN-TI achieve the top-2 recognition rate among all the variants, which confirms the visual results in Fig.~\ref{fig:4}. For poses with $45^o$, FFN-S w/o PNM, $\mathcal{L}_{\text{icd}}$, $\mathcal{L}_{\text{pcd}}$ and FRN-TI w/o $\mathcal{L}_{\text{fa}}$ even perform worse than the baseline, while FFN-S and FRN-TI achieve better than the baseline across all views. It seems the larger the head pose, the greater improvements are achieved by FFN-S and FRN-TI. For $90^o$ yaw angle, FFN-S achieves 80.11\% recognition rate, 61.31\% higher than b, while FRN-TI achieves 61.03\% higher recognition rate than the baseline.

\subsubsection{Comparison with State-of-the-Arts}
Tab.~\uppercase\expandafter{\romannumeral2} shows recognition rate comparisons of FFN-S and FRN-TI with other methods on Multi-PIE \textbf{Setting-2}. FFN-S achieves the best recognition performance across all poses among all the compared methods, followed by FRN-TI. For poses within $\pm 45^\circ$, FFN-S and FRN-TI achieve comparable recognition results with CAPG-GAN \cite{hu2018pose}. However, for poses larger than $45^\circ$, FFN-S and FRN-TI achieve much better recognition performance than CAPG-GAN. In particular, FFN-S and FRN-TI outperform CAPG-GAN by 4.15\% and 3.56\% under pose $\pm 90^\circ$, respectively.

\begin{table*}
\begin{center}
\caption{Rank-1 recognition rates (\%) across different views on Multi-PIE Setting-2. b denotes the performance of ResNet-50 on original profile images. ``-'' means the result is not reported}
\resizebox{0.6 \textwidth}{!}{
\begin{tabular}{c l c c c c c c}
\hline
1 & Methods \quad & $\pm 90^\circ$ & $\pm 75^\circ$ & $\pm 60^\circ$ & $\pm 45^\circ$  & $\pm 30^\circ$ & $\pm 15^\circ$\\
\hline
2 & b & 15.50 & 55.10 & 85.92 & 97.13 & 98.41 & 98.62 \\
\hline
3 & MVP\cite{zhu2014multi} & - & - & 60.10 & 72.90 & 83.70  & 92.80\\
4 & CPF\cite{yim2015rotating} & - & - & 61.90 & 79.90 & 88.50 & 95.00\\
5 & DR-GAN \cite{tran2017disentangled} & - & - & 83.20 & 86.20 & 90.10 & 94.00\\
6 & TP-GAN \cite{huang2017beyond} & 64.64 & 77.43 & 87.72 & 95.38 & 98.06 & 98.68\\
7 & CAPG-GAN \cite{hu2018pose}  & 66.05 & 83.05 & 90.63 & 97.33 & 99.56 & 99.82\\
\rowcolor{gray!20}
8 & FFN-S & 70.20 & 85.31 & 91.81 & 98.05 & 99.82 & 99.83\\ \rowcolor{gray!20}
9 & FRN-TI & 69.61 & 84.93 & 91.04 & 97.76 & 99.73 & 99.82\\
\hline
\end{tabular}}
\end{center}
\label{tab:2}
\end{table*}

To further validate our model's generalizability to in-the-wild face images, we qualitatively compare the visual frontalization results of FFN-S and FRN-TI with Hassner \textit{et al.} \cite{hassner2015effective}, DR-GAN \cite{tran2017disentangled}, TP-GAN \cite{huang2017beyond} and CAPG-GAN \cite{hu2018pose} on LFW datasets in Fig.~\ref{fig:5}. It is quite obvious that all the methods perform well on small pose cases. See \textit{Rows} 1 \& 2. However, for large pose cases, \textit{i.e., Rows} 3, 4 and 5, Hassner \textit{et al.}, DR-GAN, TP-GAN and CAPG-GAN distort the output faces and involve artifacts to some extent, and the face identities are also changed. FFN-S (\textit{Col.} 6) and FRN-TI (\textit{Col.} 7) still faithfully recover high-fidelity frontalized face images with finer local details and global face shapes while well preserving the identities.

Different from DR-GAN, TP-GAN and CAPG-GAN that need to provide the initial input or target facial landmarks to help face frontalization, FRN-TI does not need any facial landmark information in inference. During FRN-TI training, the normalized pose is embedded into FFN's outputs as well as the last blocks' feature maps to guide the learning of FRN, enabling FRN to spontaneously perceive the input and target poses and then map face regions to frontal.

\subsection{Evaluation on Face Restoration}
In unconstrained environment, the face image captured by a camera can be a high-quality one, which presents no degraded factors. A desired face restoration model is expected to preserve the original details for the high-quality inputs. Therefore, we first investigate the case when the inputs are high-quality face images. The visualization results of FRN-S and FRN-TI are shown in Fig.~\ref{fig:add1}. FRN-S means the face restoration net is trained separately, while FRN-TI means the restoration net is trained by the task-integrated strategy. As can be seen, the outputs of FRN-S (\textit{Row} 2) are almost the same as with the original ones, meaning FRN separately training is able to well preserve the fine details of the original high-quality face images, without any degradation of image quality. The results of FRN-TI (\textit{Row} 3) further confirm the effectiveness of TI training on face frontalization.

Then we investigate the case when the inputs are low-quality face images. We verify the effectiveness of FRN-S and FRN-TI on face restoration using two datasets, LFW and IJB-C. We also compare with FRN-FFN where face restoration and frontalization are performed separately one by one. For LFW, we generate the low-quality face images using the same methods on CASIA-Webface, denoted as LFW-Lq dataset. IJB-C contains many low-quality samples collected from real-world applications. We use NIMA \cite{talebi2018nima} to select low-quality ones for evaluation, denoted as IJBC-Lq dataset. We empirically set the confidence value $\tau$ of NIMA as 4.9 and 13, 729 image pairs are selected. As few existing methods address all the low-quality modalities at one shot, we compare our results with state-of-the-art methods that focus on each modality separately. Specifically, the super-resolution methods ESRGAN \cite{wang2018esrgan} and SI \cite{zhang2018super}, the denoising methods FFDNet \cite{zhang2018ffdnet} and PD-Denoising \cite{zhou2019awgn} as well as the deblurring method Deblur-GAN \cite{kupyn2018deblurgan} are compared with FRN-S, FRN-FFN and FRN-TI. All the methods are trained on CASIA-Webface using the same low-quality processing methods.
\begin{figure*}[!t]
\centering{\includegraphics[width = 17.5cm, height=10cm]{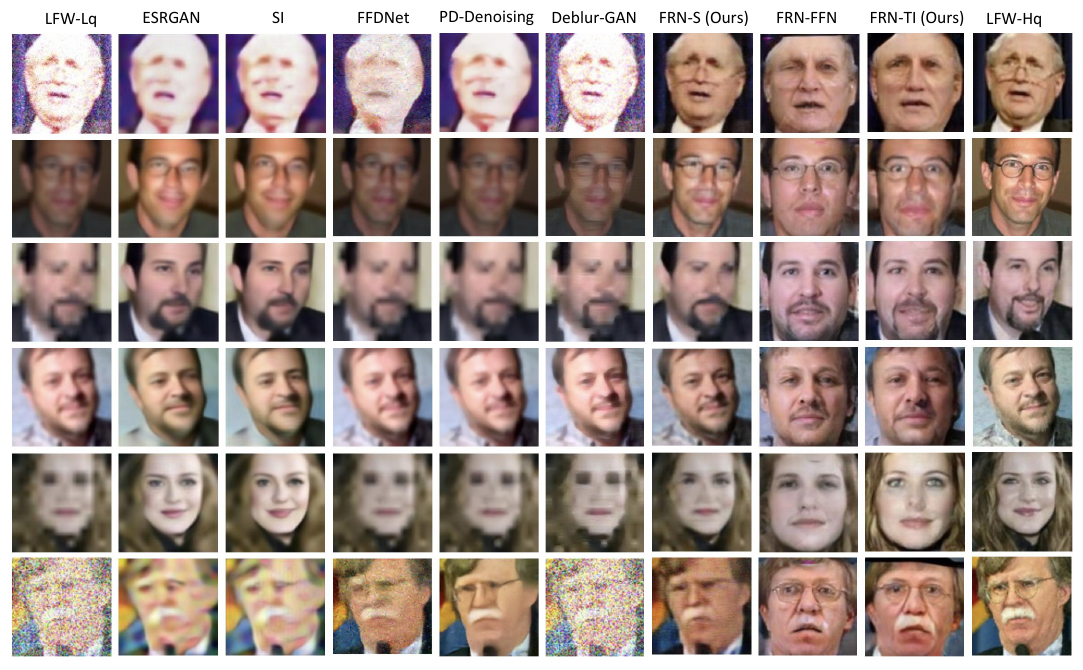}}
\vspace{-0.3cm}
\caption{Comparison of face restoration on LFW-Lq. The low-quality factors contain low resolution (\textit{Rows} 3 \& 5), bad illumination (\textit{Rows} 1 \& 2), image noise (\textit{Rows} 1 \& 6), and image blur (\textit{Rows} 2 \& 4). The last column shows the high-quality face images of LFW
\label{fig:6}}
\end{figure*}

\begin{figure}[!t]
\centering{\includegraphics[width = 8cm, height=8cm]{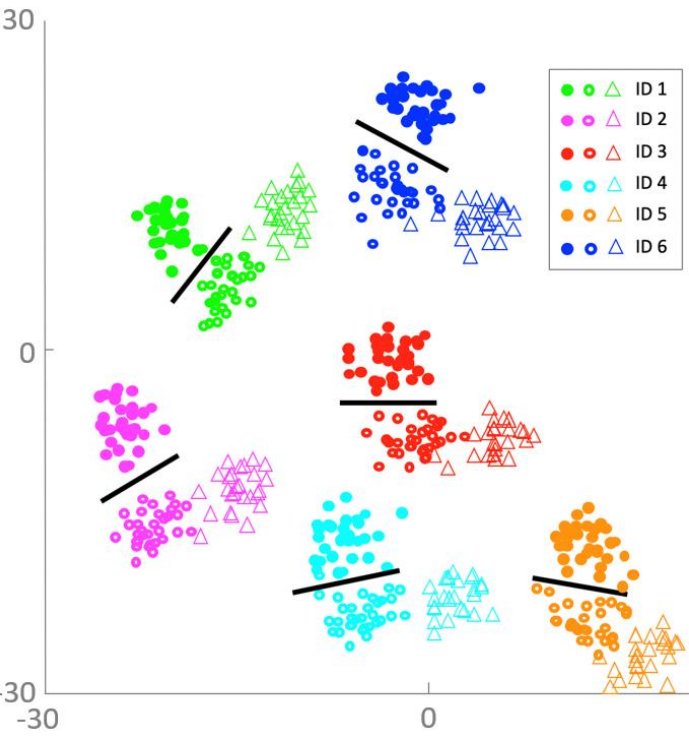}}

\caption{Distribution of identity features from FRN-S domain (dot), FFN-S domain (circle) and FRN-TI domain (triangle). The identities are randomly selected from the training set. We observe a large gap between FRN-S and FFN-S domains in the identity metric space
\label{fig:7}}
\end{figure}

\begin{figure*}[!t]
\centering{\includegraphics[width = 17.5cm, height=9.5cm]{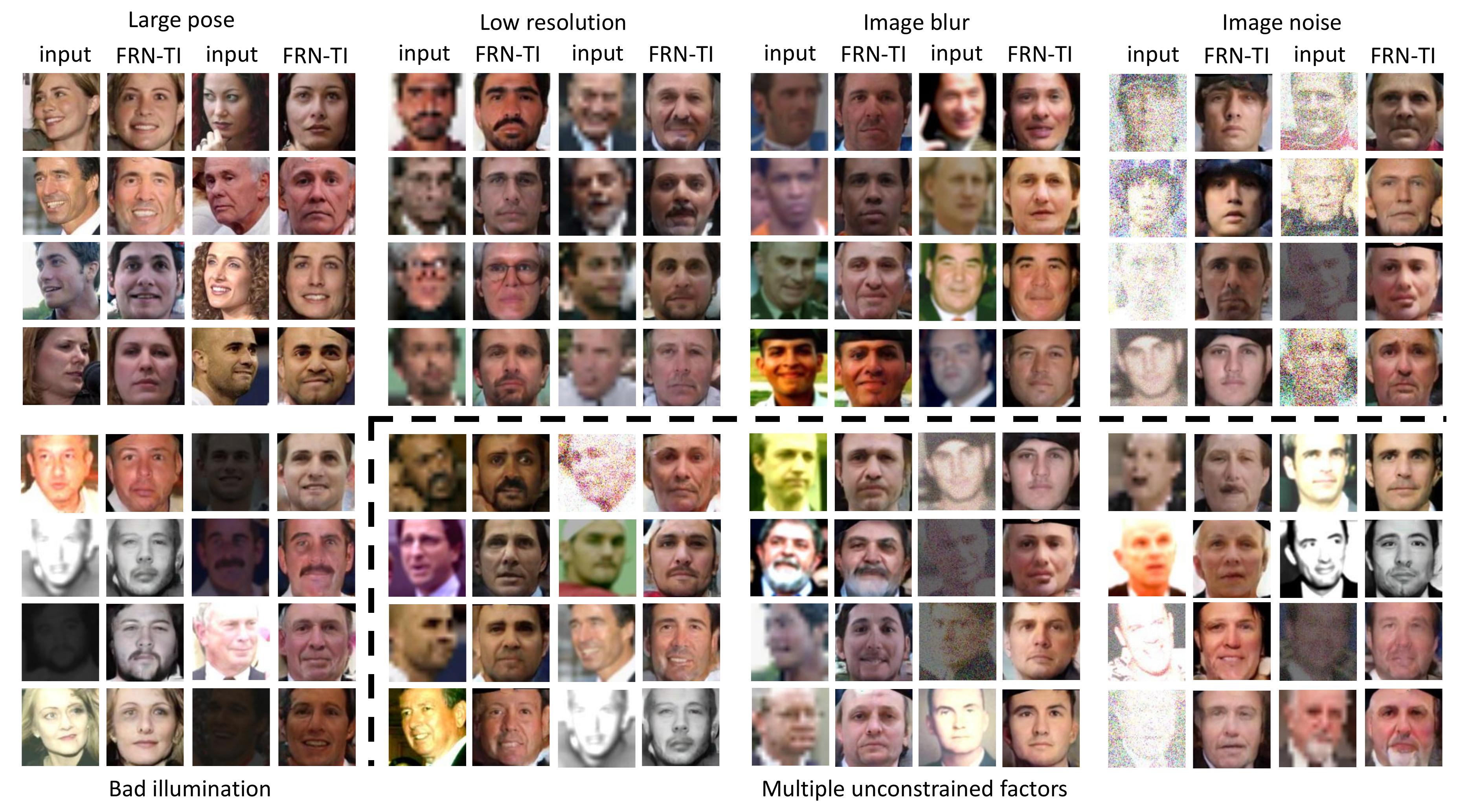}}
\vspace{-0.5cm}
\caption{Case study in extremely environments. Face images that present very large pose, very low resolution, image blur, image noise, bad illumination and multiple unconstrained factors are studied. The images are selected from LFW.
\label{fig:add2}}
\end{figure*}

We first compare the visual results of FRN-S, FRN-FFN and FRN-TI with other methods on LFW-Lq. For super-resolution, we up-sample the low-resolution inputs to $128 \times 128$ as the inputs of our models. The comparison results are shown in Fig.~\ref{fig:6}. By comparing \textit{Rows} 3 \& 5, we observe ESRGAN, SI and FRN-S are all effective to hallucinate high-resolution images from the low-resolution inputs. ESRGAN and SI achieve better visual results than FRN-S. This is because FRN-S only uses $\ell_2$ pixel-wise loss, while ESRGAN and SI use other losses like adversarial or perceptual loss to recover more facial details. When training FFN, we use the adversarial loss for supervision, thus the outputs of FRN-FFN and FRN-TI are much sharper with more high-frequency details preserved than FRN-S. Moreover, ESRGAN and SI cannot address other low-quality cases, such as bad illumination (\textit{Rows} 1 \& 2), noise (\textit{Rows} 1 \& 6) and image blur (\textit{Rows} 2 \& 4), while FRN-S, FRN-FFN and FRN-TI are effective for all low-quality factors. Similar experimental results can also be obtained from denoising methods FFDNet and PD-Denoising, and also deblurring method Deblur-GAN. FFDNet and PD-Denoising can only address the problem of image noise (\textit{Rows} 1 \& 6), but they fail on others such as low resolution, bad illumination and image blur. For Deblur-GAN, it is only effective on blurred images (\textit{Rows} 2 \& 4) and fails on others. FRN-FFN and FRN-TI recover more facial details than FRN-S and meanwhile frontalize the profile face to its frontal pose. Compared with FRN-IT, FRN-FFN loses more facial details and slightly changes the face identity in some cases (\textit{Rows} 2 \& \textit{5}). The reason behind this may be the identity representation inconsistency between the outputs of FRN-S and FFN-S. To investigate this challenge, we visualize the learned identity features in Fig.~\ref{fig:7} and observe there exists a large margin between FRN-S domain and FFN-S domain. This identity inconsistency presents a gap when performing face frontalization from the outputs of FRN-S, which makes the model difficult to converge and affects the identity preservation.

From the results of Fig.~\ref{fig:add1} and Fig.~\ref{fig:6}, we can conclude that the proposed FRN is able to output high-quality face images,  no matter the inputs are degenerated face images or high-quality ones. Since we overlay different degradation factors such as low-resolution, noise, blur and bad illumination during data processing, our restoration task is more challenging than the traditional ones that focus on only one degradation factor. However, there are two limitations to the proposed FRN. The first one is that, our training data is driven by artificial degradation method, which may reduce the efficacy when addressing unseen data from real world. The second one is that, our method only uses the high-quality images to guide feature learning without considering the entangled influence of difference degradation factor in feature space, which may reduce the representation ability for the encoded features. Perhaps the disentangled representation learning is a better solution, which can remove the influence of other factors, and choose the best one for face restoration.

We follow the literatures \cite{yu2019can,zhang2021face} for a qualitative comparisons with other state-of-the-art methods. For a fair comparison, we only consider the low-resolution factor during the retraining of our model. Since the works \cite{yu2019can,zhang2021face} also focus on joint face frontalization and restoration and have reported their results, we copy the reported results from the work \cite{zhang2021face} and compare with ours on the dataset Multi-PIE and CelebA \cite{liu2015deep}, using the average Peak Signal-to-Noise (PSNR) and the Structural SIMilarity (SSIM) scores. The qualitative comparison results are reported in Tab.~\uppercase\expandafter{\romannumeral3}. As can be seen, the proposed MDFR achieves the best performance in both metrics on Multi-PIE and CelebA datasets. Specifically, MDFR outperforms VividGAN by 0.967 dB and 0.882 dB in PSNR on the datasets Multi-PIE and CelebA, respectively. For the metric SSIM, MDFR outperforms VividGAN by 0.022 and 0.014 on Multi-PIE and CelebA, respectively. The results indicate that the proposed MDFR achieves more authentic results than the other comparison methods.

\begin{table}[t!]
\small
\begin{center}
\caption{Quantitative comparison results with other methods.}
\resizebox{0.45 \textwidth}{!}{
\begin{tabular}{ c c c c c c }
\hline
1 \quad &  & \multicolumn{2}{c}{Multi-PIE} & \multicolumn{2}{c}{CelebA}  \\
\hline
2 & & PSNR & SSIM & PSNR & SSIM  \\
\hline
3 & TANN \cite{yu2019can} & 24.426 & 0.831 & 25.690 & 0.870     \\
4 & VividGAN \cite{zhang2021face} & 26.289 & 0.876 & 26.965 & 0.893    \\
5 & MDFR (ours) & \textbf{27.256} & \textbf{0.898} & \textbf{27.847} & \textbf{0.907}    \\
\hline
\end{tabular}}
\end{center}
\label{tab:add}
\end{table}

We then report face recognition performance of FRN-S, FRN-FFN and FRN-TI and compare with other state-of-the-art face restoration methods on LFW-Lq and IJBC-Lq. Specifically, we use face restoration methods to recover high-quality images from low-quality ones and then use Resnet-50 to extract identity features for face recognition. The recognition results are reported in Tab.~\uppercase\expandafter{\romannumeral4}, where b is the result on low-quality images without restoration used for face recognition directly. From Tab.~\uppercase\expandafter{\romannumeral4}, we find FRN-TI achieves the best performance on both LFW-Lq and IJBC-Lq, followed by FRN-FFN and FRN-S. Compared with b, FRN-TI improves face recognition rate by 23.61\% and 9.93\% on LFW-Lq and IJBC-Lq, respectively; FRN-FFN improves by 22.29\% and 7.88\%, respectively; FRN-S improves by 21.89\% and 6.17\%, respectively. By comparing FRN-S and FRN-TI, FRN-TI's outputs contain more facial details which can be used for recognition. In addition, the enhanced faces are normalized to frontal, further boosting the performance for FRN-TI than FRN-S. As FRN-FFN performs high-quality face frontalization in two separate phases, the identity representation gap between these two tasks hinders them from linking well to each other, hence decreasing the final recognition results. As other compared methods can only address a certain low-quality factor, they report poor results on LFW-Lq. Even-though they are effective for a specific low-quality aspect, they may fail on others. In addition, no significant recognition improvements are observed using the compared methods on IJBC-Lq, which contains real world low-quality samples. Some of the methods, \textit{e.g.,} ESRGAN and FFDNet even slightly decrease the recognition performance compared with the baseline.

We then use our MDFR (FRN-TI) as face image preprocessing and use the face recognition methods, Sphereface \cite{liu2017sphereface}, CosFace \cite{wang2018cosface} and ArcFace \cite{deng2019arcface} to perform face verification. The comparison results are shown in Tab.~\uppercase\expandafter{\romannumeral5}. It is clear to see, when MDFR is used to pre-process the original face images, the recognition performance for each of the method is improved. Comparing with the state-of-the-art face recognition method ArcFace, if MDFR is used the recognition can be improved by 0.03\%, 0.05\% and 0.06\% on LFW, AgeDB-30 and CFP-FP, respectively. Based on the observation, our ARFM can serve as a plug-and-play module to any state-of-the-art methods for high-performance unconstrained face recognition.

\begin{table*}[t!]
\small
\begin{center}
\caption{Quantitative comparisons on identity recognizability. b denotes the results on low-quality images without restoration. \textit{Row} 3 reports the face verification acc (\%) on LFW-Lq; \textit{Row} 4 reports the face verification acc (\%) on IJBC-Lq @FFR = 0.1}

\resizebox{1.0 \textwidth}{!}{
\begin{tabular}{ c c c c c c c c c c}
\hline
Methods \quad & b & ESRGAN  & SI  & FFDNet  & PD-Denoising  & Deblur-GAN  & FRN-S & FRN-FFN & FRN-TI  \\
\hline
Input Size & 128$\times$128 & 12$\times$14 & 12$\times$14 & 128$\times$128 & 128$\times$128 & 128$\times$128 & 128$\times$128 & 128$\times$128 & 128$\times$128 \\
LFW-Lq Acc. & 71.62 & 78.21 & 83.50 & 75.21 & 78.35 & 76.32 & 93.51 & 93.91 & \textbf{95.23}    \\
IJBC-Lq Acc. & 64.25 & 63.25 & 67.23 & 64.21 & 64.26 & 65.12 & 70.42 & 72.13 & \textbf{74.18}   \\
\hline
\end{tabular}}
\end{center}
\label{tab:3}
\end{table*}

\begin{table*}[t!]
\small
\begin{center}
\caption{Face verification results (\%) of different methods on LFW, AgeDB-30 \cite{moschoglou2017agedb} and CFP-FP \cite{sengupta2016frontal}. MDFR serves as pre-processing of face images for each of the methods}

\resizebox{0.93 \textwidth}{!}{
\begin{tabular}{ c c c c c c c}
\hline
Methods \quad & SphereFace & SphereFace + MDFR & CosFace & CosFace + MDFR & ArcFace & ArcFace + MDFR  \\
\hline
LFW \cite{huang2008labeled} & 99.42 & 99.48 & 99.51 & 99.54 & 99.53 & \textbf{99.56}    \\
AgeDB-30 \cite{moschoglou2017agedb} & 91.70 & 91.81 & 94.56 & 94.68 & 95.15 & \textbf{95.20}  \\
CFP-FP \cite{sengupta2016frontal} & 94.38 & 94.51 & 95.44 & 95.59 & 95.56 & \textbf{95.62}  \\
\hline
\end{tabular}}
\end{center}
\label{tab:4}
\end{table*}

For more visualization results of FRN-TI, please refer to Fig.~\ref{fig:add2}, where we have shown case study results in extremely environments, \textit{i.e.}, very large pose ($1^{st}$ panel), very low resolution ($2^{nd}$ panel), image blur ($3^{rd}$ panel), image noise ($4^{th}$ panel), bad illumination ($5^{th}$ panel) and multiple contamination factors ($6^{th}-8^{th}$ panels) that contain at least two of the unconstrained factors. It is clear to see that our FRN-TI is able to generate photorealistic and identity-preserving frontal view faces from profile face images, no matter the profile images contain only one unconstrained factor or multiple unconstrained factors. In particular, when the input face images present low quality factors such as low resolution, noise, blur or bad illumination that may fail facial landmark detection, our model still can recover the high-quality and frontal view counterpart, which is more advanced than many of the landmark guided face frontalization methods.

\section{Conclusion}

In this paper, we propose a novel \textbf{M}ulti-\textbf{D}egradation \textbf{F}ace \textbf{R}estoration (MDFR) model. MDFR contains a dual-agent generator and a dual-agent prior-guided discriminator which cooperate with each other to learn frontalized high-quality faces from face images with multiple low-quality factors and arbitrary facial poses. The Pose Normalization Module (PNM) based on a 3D morphable model is proposed to normalize facial landmarks to real-frontal, serving as a unified criterion to guide the learning of MDFR. The Task-Integrated (TI) training is further developed to merge face restoration and frontalization into a unified network. When the TI training is done, MDFR is able to restore frontal and high-quality face images from low-quality ones with arbitrary facial pose without requesting any prior input landmarks. We demonstrate that the proposed unified framework outputs more visually realistic face images with more discriminative features preserved for face recognition than performing face restoration and frontalization separately. Comprehensive experiments on both controlled and ``in-the-wil'' face benchmarks illustrate the superiority of our method compared with other state-of-the-art face frontalization and face restoration methods.

\section*{Acknowledgment}
This work was partially supported by the Open Fund Project of Key Laboratory of Flight Technology and Flight Safety of Civil Aviation (FZ2020KF10), the National Science Foundation of China (62006244), the Project of Comprehensive Reform of Electronic Information Engineering Specialty for Civil Aircraft Maintenance (14002600100017J172), the Project of Civil Aviation Flight University of China (Grant Nos. J2018-56,  CJ2019-03, J2020-060), and the Sichuan University Failure Mechanics \& Engineering Disaster Prevention and Mitigation Key Laboratory of Sichuan Province Open Foundation (2020FMSCU02).

\begin{IEEEbiography}[{\includegraphics[width=1.1in,height=1.35in,clip,keepaspectratio]{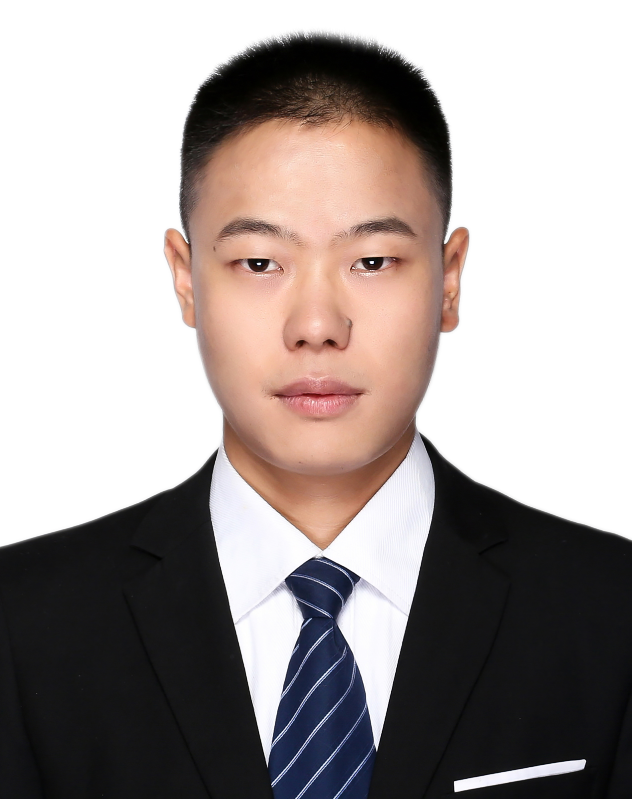}}]%
{Xiaoguang Tu}
is currently a lecturer in Aviation Engineering Institute at Civil Aviation Flight University of China. He received his Ph.D degree from the University of Electronic Science and Technology of China (UESTC) in 2020. He was a visiting scholar at Learning and Vision Lab, National University of Singapore (NUS) from 2018 to 2020 under the supervision of Dr. Jiashi Feng. His research interests include convex optimization, computer vision and deep learning.
\end{IEEEbiography}

\vspace{-0.6cm}

\begin{IEEEbiography}[{\includegraphics[width=1.1in,height=1.35in,clip,keepaspectratio]{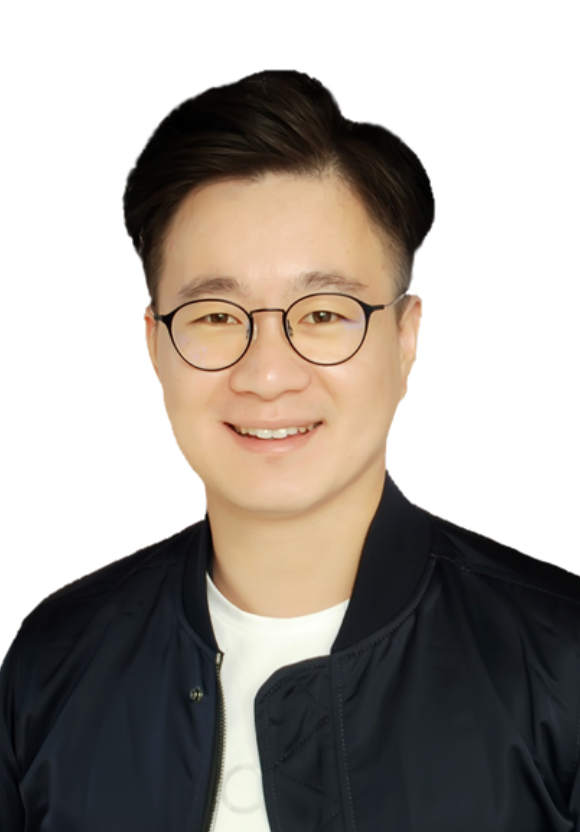}}]%
{Jian Zhao}
received the Bachelor degree from Beihang University in 2012, the Master degree from the National University of Defense Technology in 2014, and the Ph.D. degree from the National University of Singapore in 2019. He is currently an Assistant Professor with the Institute of North Electronic Equipment, Beijing, China. His main research interests include deep learning, pattern recognition, computer vision, and multimedia analysis. He has published over 40 cutting-edge papers. He has received the Young Talent Support Project from China Association for Science and Technology, and Beijing Young Talent Support Project from Beijing Association for Science and Technology, the Lee Hwee Kuan Award (Gold Award) on PREMIA 2019, the Best Student Paper Award on ACM MM 2018, and the top-3 awards several times on worldwide competitions. He is the SAC of VALSE, and the committee member of CSIG-BVD. He has served as the invited reviewer of NSFC, T-PAMI, IJCV, NeurIPS (one of the top 30\% highest-scoring reviewers of NeurIPS 2018), CVPR, etc.
\end{IEEEbiography}

\vspace{-0.5cm}

\begin{IEEEbiography}[{\includegraphics[width=1.1in,height=1.35in,clip,keepaspectratio]{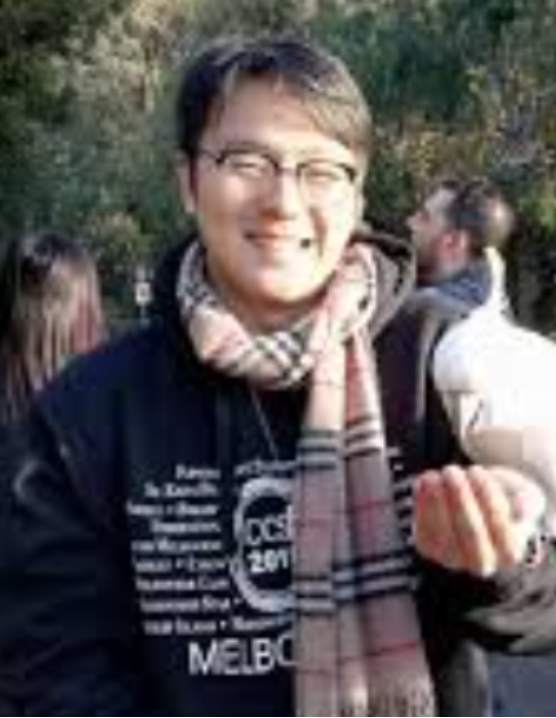}}]%
{Qiankun Liu} is an AI scientist at Pensees Pte. Ltd. Singapore. His research interests include generative adversarial networks, optical flow estimation and face recognition.
\end{IEEEbiography}

\begin{IEEEbiography}[{\includegraphics[width=1.1in,height=1.35in,clip,keepaspectratio]{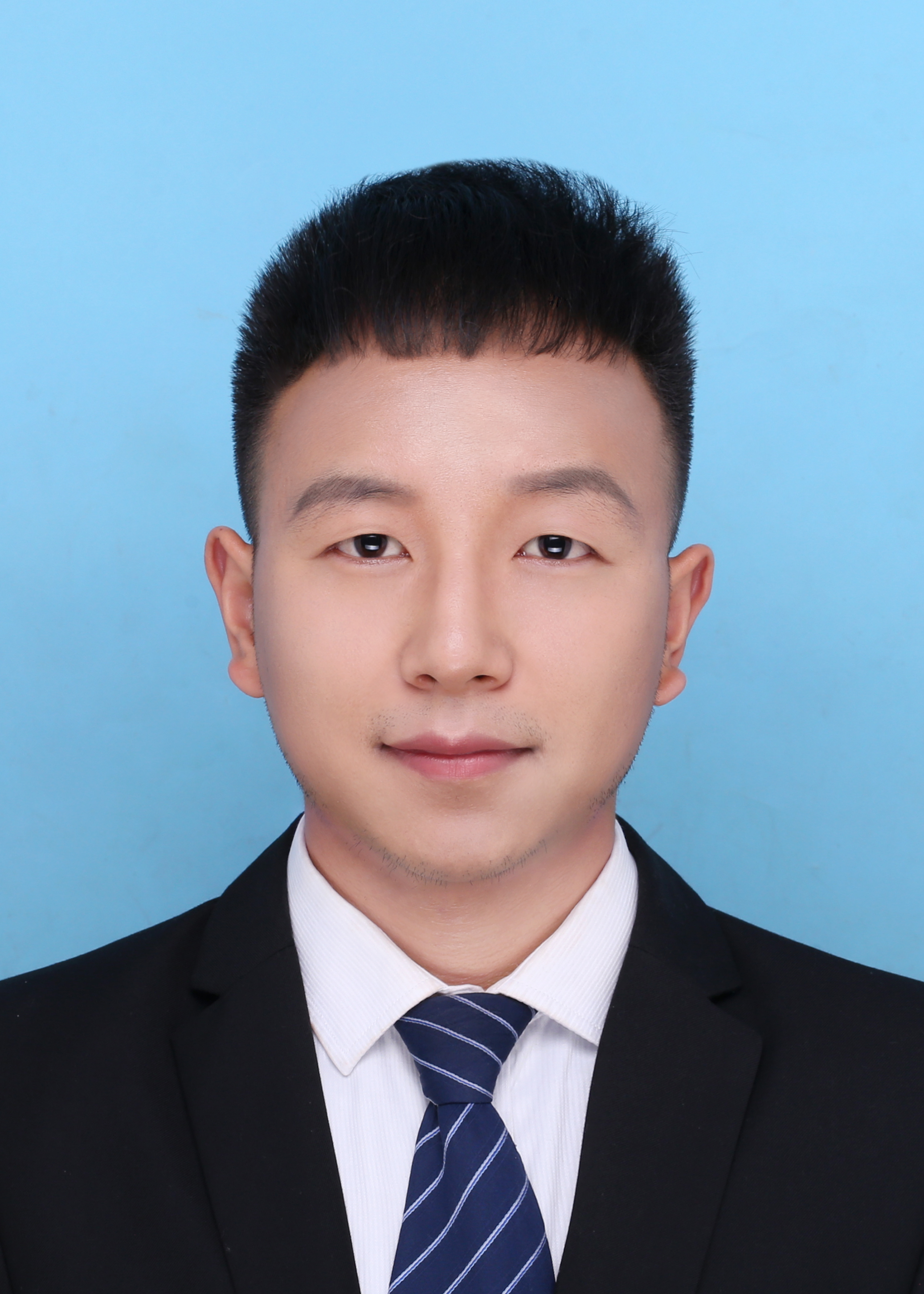}}]%
{Wenjie Ai}
is a Master with the School of Information and Communication Engineering at University of Electronic Science and Technology of China (UESTC). His research areas of interest mainly include computer vision and deep learning, in particular, super resolution and deblur.
\end{IEEEbiography}

\vspace{-0.5cm}

\begin{IEEEbiography}[{\includegraphics[width=1.1in,height=1.35in,clip,keepaspectratio]{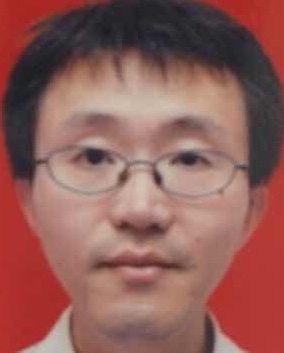}}]%
{Zhifeng Li}
is currently a top-tier principal researcher with Tencent AI Lab. He received the Ph. D. degree from the Chinese University of Hong Kong in 2006. After that, He was a postdoctoral fellow at the Chinese University of Hong Kong and Michigan State University for several years. Before joining Tencent AI Lab, he was a full professor with the Shenzhen Institutes of Advanced Technology, Chinese Academy of Sciences. His research interests include deep learning, computer vision and pattern recognition, and face detection and recognition. He is currently serving on the Editorial Boards of Neurocomputing and IEEE Transactions on Circuits and Systems for Video Technology. He is a fellow of British Computer Society (FBCS).
\end{IEEEbiography}

\vspace{-0.5cm}

\begin{IEEEbiography}[{\includegraphics[width=1.1in,height=1.35in,clip,keepaspectratio]{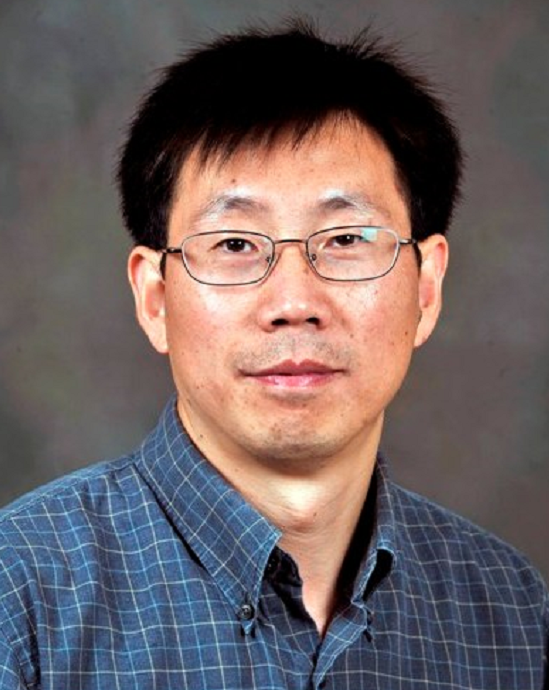}}]%
{Guodong Guo}
received the Ph.D. degree in computer science from University of Wisconsin, Madison, WI, USA. He is currently the Deputy Head of the Institute of Deep Learning, Baidu Research, and also an Associate Professor with the Department of Computer Science and Electrical Engineering, West Virginia University (WVU), USA. His research interests include computer vision, biometrics, machine learning, and multimedia. He received the North Carolina State Award for Excellence in Innovation in 2008, Outstanding Researcher (2017-2018, 2013-2014) at CEMR, WVU, and New Researcher of the Year (2010-2011) at CEMR, WVU. He was selected the ``People's Hero of the Week'' by BSJB under Minority Media and Telecommunications Council (MMTC) in 2013. Two of his papers were selected as ``The Best of FG'13" and ``The Best of FG'15", respectively.
\end{IEEEbiography}

\vspace{-0.5cm}

\begin{IEEEbiography}[{\includegraphics[width=1.1in,height=1.35in,clip,keepaspectratio]{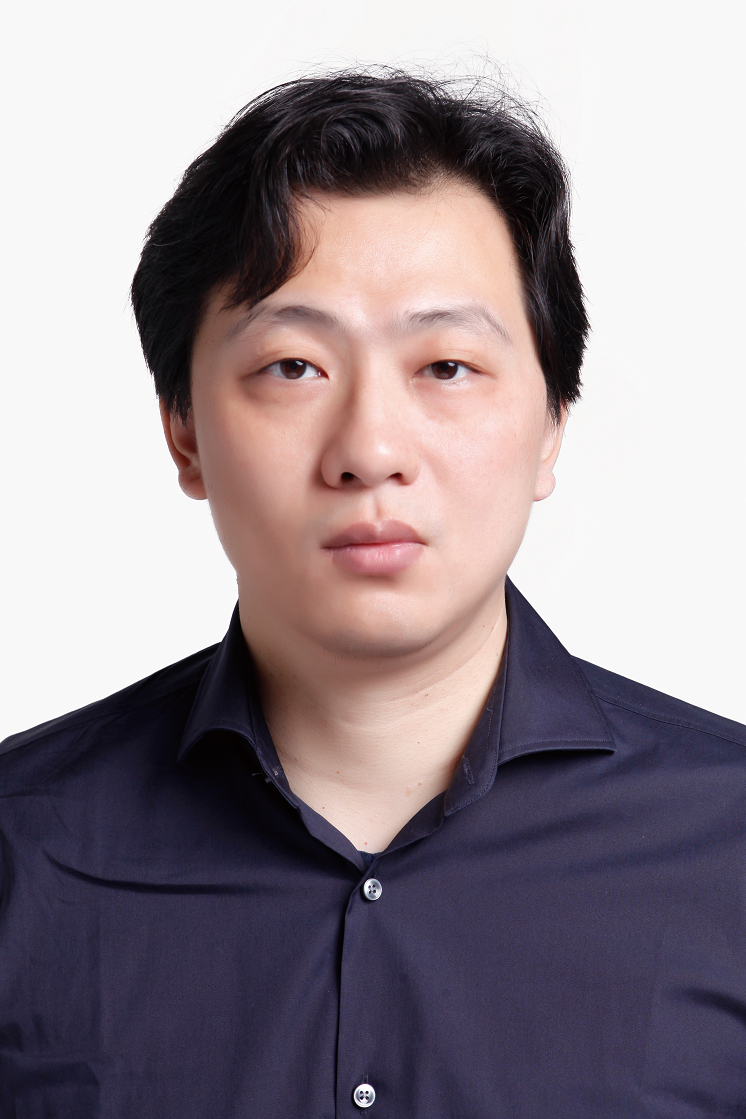}}]%
{Wei Liu}
is currently a Distinguished Scientist of Tencent, China and a director of Computer Vision Center at Tencent AI Lab.  Prior to that, he has been a research staff member of IBM T. J. Watson Research Center, Yorktown Heights, NY, USA from 2012 to 2015. Dr. Liu has long been devoted to research and development in the fields of machine learning, computer vision, pattern recognition, information retrieval, big data, etc. Dr. Liu currently serves on the editorial boards of IEEE Transactions on Pattern Analysis and Machine Intelligence, IEEE Transactions on Neural Networks and Learning Systems, IEEE Transactions on Circuits and Systems for Video Technology, Pattern Recognition, etc. He is a Fellow of the International Association for Pattern Recognition (IAPR) and an Elected Member of the International Statistical Institute (ISI).
\end{IEEEbiography}

\vspace{-0.5cm}

\begin{IEEEbiography}[{\includegraphics[width=1.1in,height=1.35in,clip,keepaspectratio]{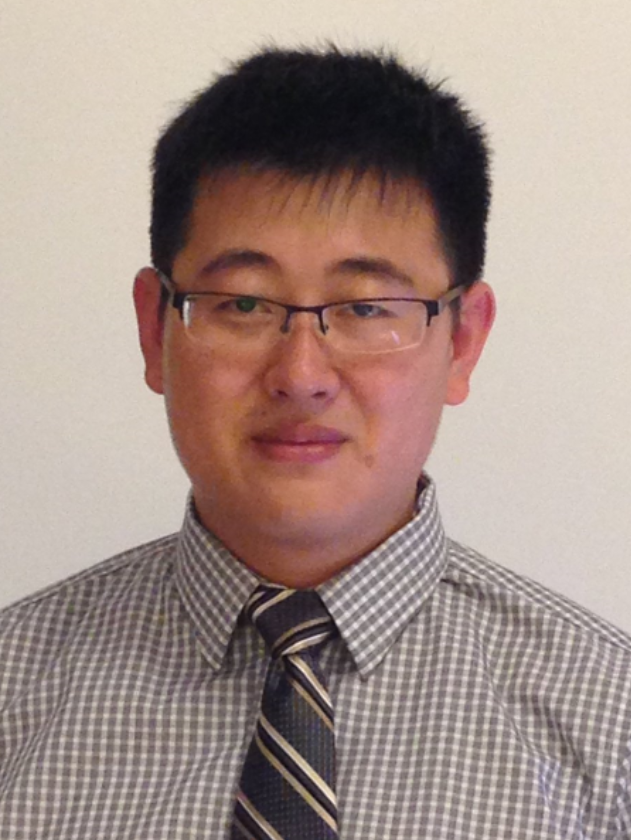}}]%
{Jiashi Feng}
received the B.E. degree from University of Science and Technology of China, Hefei, China, in 2007, and the Ph.D. degree from the National University of Singapore, Singapore, in 2014. He was a Post-Doctoral Researcher with the University of California from 2014 to 2015. He is currently an Assistant Professor with the Department of Electrical and Computer Engineering, National University of Singapore. His current research interests focus on machine learning and computer vision techniques for large-scale data analysis.
\end{IEEEbiography}

\end{document}